\documentclass[letterpaper]{article} 
\usepackage{aaai23}  
\usepackage{times}  
\usepackage{helvet}  
\usepackage{courier}  
\usepackage[hyphens]{url}  
\usepackage{graphicx} 
\usepackage{natbib}  
\usepackage{caption} 
\usepackage{algorithm}
\usepackage{algorithmic}
\usepackage{mathrsfs}
\usepackage{multirow}
\usepackage{xspace}
\usepackage{subfigure} 
\usepackage{threeparttable}
\usepackage{amsmath}
\usepackage{amssymb}
\usepackage{mathrsfs}
\usepackage{amsfonts}
\usepackage{xcolor}
\usepackage{epsfig}
\usepackage{diagbox}
\usepackage{multirow}
\usepackage{booktabs}
\usepackage{bm}
\usepackage{amssymb}
\usepackage{float}
\newcommand{\R}{\mathbb{R}}
\usepackage{newfloat}
\usepackage{listings}
\DeclareCaptionStyle{ruled}{labelfont=normalfont,labelsep=colon,strut=off} 
\floatstyle{ruled}
\newfloat{listing}{tb}{lst}{}
\floatname{listing}{Listing}
\title{Dish-TS: A General Paradigm for Alleviating Distribution Shift in \\ Time Series Forecasting}

\author {
    Wei Fan,\textsuperscript{\rm 1}\thanks{Partial of the work is done at IOTSC, University of Macau.}
    Pengyang Wang,\textsuperscript{\rm 2}\thanks{Corresponding authors.}
    Dongkun Wang,\textsuperscript{\rm 2}
    Dongjie Wang,\textsuperscript{\rm 1}
    Yuanchun Zhou,\textsuperscript{\rm 3}
    Yanjie Fu\textsuperscript{\rm 1}\footnotemark[2]
}
\affiliations {
    \textsuperscript{\rm 1}Department of Computer Science, University of Central Florida, \\
    \textsuperscript{\rm 2}State Key Laboratory of Internet of Things for Smart City, University of Macau, \\
    \textsuperscript{\rm 3}Computer Network Information Center, Chinese Academy of Sciences \\
    \{weifan, wangdongjie\}@knights.ucf.edu,  \{pywang, yc17908\}@um.edu.mo, 
    zyc@cnic.cn, yanjie.fu@ucf.edu
}

\begin{document}

\maketitle

\begin{abstract}
The distribution shift in Time Series Forecasting (TSF), indicating series distribution changes over time, largely hinders the performance of TSF models.
Existing works towards distribution shift in time series are mostly limited in the quantification of distribution and, more importantly, overlook the potential shift between lookback and horizon windows. To address above challenges, we systematically summarize the distribution shift in TSF into two categories.
Regarding lookback windows as input-space and horizon windows as output-space, there exist (i) {\it intra-space shift}, that the distribution within the input-space keeps shifted over time, and
(ii) {\it inter-space shift}, that the distribution is shifted between input-space and output-space.
Then we introduce, \textit{Dish-TS}, a general neural paradigm for alleviating distribution shift in TSF. 
Specifically, for better distribution estimation, we propose the coefficient net (\textsc{Conet}), which can be any neural architectures, to map input sequences into learnable distribution coefficients. 
To relieve {\it intra-space} and {\it inter-space shift}, we organize \textit{Dish-TS} as a Dual-\textsc{Conet} framework to separately learn the distribution of input- and output-space, which naturally captures the distribution difference of two spaces. In addition, we introduce a more effective training strategy for intractable {Conet} learning.
Finally, we conduct extensive experiments on several datasets coupled with different state-of-the-art forecasting models. Experimental results show \textit{Dish-TS} consistently boosts them with a more than 20\% average improvement. Code is available at \texttt{{https://github.com/weifantt/Dish-TS}}.
\end{abstract}

\section{Introduction} \label{sec:intro}

Time Series Forecasting (TSF) has been playing an essential role in many applications, such as electricity consumption planning~\cite{akay2007grey}, transportation traffic analysis~\cite{ming2022multi}, weather condition estimation \cite{abhishek2012weather,han2021joint}.  Following by traditional statistical methods, (\textit{e.g.,} \cite{holt1957forecasting}), deep learning-based TSF models, (\textit{e.g.,}, \cite{salinas2020deepar,rangapuram2018deep}), have achieved great performance in various areas. 

Despite the remarkable success of TSF models, the non-stationarity of time series data has been an under-addressed challenge for accurate forecasting~\cite{hyndman2018forecasting}. The non-stationarity, depicting the distribution of series data is shifted over time, can be interpreted as distribution shift in time series~\cite{brockwell2009time}. Such problem results into poor model generalization, thus largely hindering the performance of time series forecasting.

\begin{figure}[!t]
\centering
\includegraphics[width=0.85\linewidth]{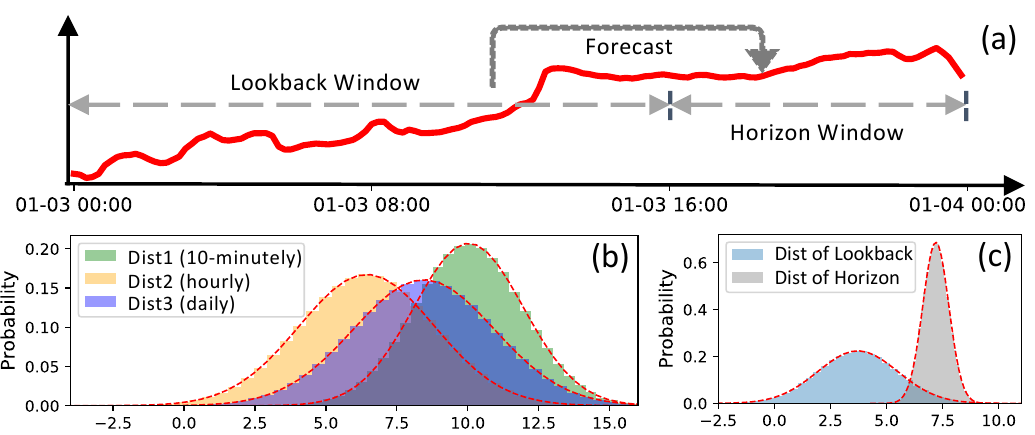}
\caption{(a) given time series (weather) data, take lookback windows to forecast horizon windows; (b) shows distributions (depicted by mean and std.) on different sampling frequencies towards one series; (c) shows the distribution difference of the lookback window and horizon window of (a).}
\label{fig:motivation}
\vspace{-3mm}
\end{figure}

After analyzing numerous series data, we systematically organize distribution shift of TSF into two categories. Considering the lookback windows (`lookbacks' for brevity) as input-space of models and horizon windows (`horizons' for brevity) as output-space of models\footnote{In this paper, we use `lookbacks/horizons', `lookback/horizon windows',`input/output-space' interchangeably.}, there are (i) \textbf{\textit{intra-space shift}}: time series distribution changes over time, making data within input-space (lookbacks) shifted;
(ii) \textbf{\textit{inter-space shift}}: the distribution is shifted between input-space (lookbacks) and output-space (horizons). 
Existing works have tried to alleviate distribution shift problem in TSF \cite{ogasawara2010adaptive, passalis2019deep, du2021adarnn, kim2022reversible}.
However, most of them exhibit two limitations:

\textbf{\textit{{First, the distribution quantification for {intra-space} in TSF is} {unreliable}.}}
Time series is ideally generated continuously from the true distribution, while the observational data is actually sampled discretely with senors in a certain recording frequency. Existing works always directly normalize or rescale the series \cite{ogasawara2010adaptive, passalis2019deep, kim2022reversible}, by quantifying true distribution with fixed statistics ({\it e.g.}, mean and std.) empirically obtained from observational data, and then normalizing series distribution with these statistics.
However, the empirical statistics are unreliable and limited in expressiveness for representing the true distribution behind the data.
For example, Figure \ref{fig:motivation}(b) indicates three distributions (depicted by mean and std.) sampled from the same series with different frequencies ({\it i.e.}, ten-minutely, hourly, daily). 
Despite coming from the same series, different sampling frequencies provide different statistics, prompting the question: which one best represents the true distribution? Since the recording frequency of time series is determined by sensors, it is difficult to identify what the true distribution is behind the data. 
Thus, how to properly quantify the distribution, as well as the distribution shift of intra-space, still remains a problem.

\textbf{\textit{Second, the \textbf{inter-space shift} of TSF is  neglected.}}
In time series forecasting, considering the input-sequences (lookbacks) and output-sequences (horizons) as two spaces, existing works always assume the input-space and output-space follow the same distribution by default~\cite{ogasawara2010adaptive,passalis2019deep,du2021adarnn}. 
Though a more recent study, RevIN \cite{kim2022reversible}, tries to align instances through normalizing the input and denormalizing the output, it still puts a strong assumption that the lookbacks and horizons share the same statistical properties; hence the same distribution. 
Nonetheless, there is always a variation in distribution between input-space and output-space. As shown in Figure \ref{fig:motivation}(c), the distribution (depicted by mean and std.) between the lookback window and the horizon window exhibits a considerable disparity. The ignorance of inter-space shift overlooks the gap between input- and output-space, thus hindering forecasting performance.

To overcome the above limitations, we propose an effective general neural paradigm, \textit{Dish-TS}, against \textit{Di}stribution \textit{sh}ift in \textit{T}ime \textit{S}eries.
\textit{Dish-TS} is model-agnostic and can be coupled with any deep TSF models. 
Inspired by \cite{kim2022reversible}, \textit{Dish-TS} includes a two-stage process, which normalizes model input before forecasting and denormalizes model output after forecasting.
To solve the problem of unreliable distribution quantification, we first propose a coefficient net (\textsc{Conet}) to measure the series distribution. Given any window of series data, \textsc{Conet} maps it into two learnable coefficients: a level coefficient and a scaling coefficient to illustrate series overall scale and fluctuation. In general, \textsc{Conet} can be designed as any neural architectures to conduct any linear/nonlinear mappings, providing sufficient modeling capacity of varied complexities.
To relieve the aforementioned \textbf{\textit{intra-space shift}} and  \textbf{\textit{inter-space shift}}, we organize \textit{Dish-TS} as a Dual-\textsc{Conet} framework. 
Specifically, Dual-\textsc{Conet} consists of two separate \textsc{Conet}s:
(1) \textsc{BackConet}, that produces coefficients to estimate the distribution of input-space (lookbacks), 
and (2) the \textsc{HoriConet}, that generates coefficients to infer the distribution of output-space (horizons).
The Dual-\textsc{Conet} setting captures distinct distributions for input- and output-space respectively, which naturally relieves the inter-space shift. 

In addition, \textit{Dish-TS} is further introduced with an effective prior-knowledge induced training strategy for \textsc{Conet} learning, considering \textsc{HoriConet} needs to infer (or predict) distribution of output-space, which is more intractable due to inter-space shift. Thus, some extra distribution characteristics of output-space is used to provide \textsc{HoriConet} more supervision of prior knowledge.
In summary, our contributions are listed:
\begin{itemize}

\item We systematically organize distribution shift in time series forecasting as {\it intra-space shift} and {\it inter-space shift}. 

\item We propose \textit{Dish-TS}, a general neural paradigm for alleviating distribution shift in TSF, built upon Dual-\textsc{Conet} with jointly considering {\it intra-space} and {\it inter-space shift}. 

\item To implement \textit{Dish-TS}, we provide a most simple and intuitive instance of \textsc{Conet} design with a prior knowledge-induced training fashion to demonstrate the effectiveness of this paradigm.

\item Extensive experiments over various datasets have shown our proposed \textit{Dish-TS} consistently boost current SOTA models with an average improvement of 28.6\% in univariate forecasting and 21.9\% in multivariate forecasting.
\end{itemize}

\section{Related Work}

\subsubsection{Models for Time Series Forecasting.}
Time series forecasting (TSF) is a longstanding research topic. At an early stage, researchers have proposed statistical modeling approaches, such as exponential smoothing \cite{holt1957forecasting} and auto-regressive moving averages (ARMA) \cite{whittle1963prediction}.
Then, more works propose more complicated models: Some researchers adopt a hybrid design \cite{montero2020fforma,smyl2020hybrid}.
With the great successes of deep learning, many deep learning models have been developed for time series forecasting \cite{rangapuram2018deep, salinas2020deepar, zia2020residual,cao2020spectral,fan2022depts}. 
Among them, one most representative one is N-BEATS that \cite{Oreshkin2020N-BEATS} applies pure fully connected works and achieves superior performance. Transformer \cite{vaswani2017attention} has been also used for series modeling. To improve it, Informer \cite{zhou2021informer} improves in attention computation, memory consumption and inference speed. Recently, Autoformer \cite{xu2021autoformer}  
replace attention with auto-correlation to facilitate forecasting.

\subsubsection{Distribution Shift in Time Series Forecasting.}
Despite of many remarkable models, time series forecasting still suffers from distribution shift considering distribution of real-world series is changing over time \cite{akay2007grey}. To solve this problem, some normalization techniques are proposed: Adaptive Norm \cite{ogasawara2010adaptive} puts z-score normalization on series by the computed global statistics. Then, DAIN \cite{passalis2019deep} applies nonlinear neural networks to adaptively normalize the series. \cite{du2021adarnn} proposed Adaptive RNNs to handle the distribution shift in time series.
Recently, RevIN \cite{kim2022reversible} proposes an instance normalization to reduce series shift.
Though DAIN has used simple neural networks for normalization, most works \cite{ogasawara2010adaptive, du2021adarnn,kim2022reversible} still used static statistics or distance function to describe distribution and normalize series, which is limited in expressiveness. Some other works study time series distribution shift in certain domains such as trading markets \cite{cao2022a}.
They hardly consider the inter-space shift between model input-space and output-space.

\section{Problem Formulations}

\paragraph{Time Series Forecasting.}
Let $x_t$ denote the value of a regularly sampled time series at time-step $t$, and the classic time series forecasting formulation is to project historical observations $\bm{x}_{t-L:t} = [x_{t-L+1}, \cdots, x_{t} ]$ into their subsequent future values $\bm{x}_{t:t+H} =  [x_{t+1}, \cdots, x_{t+H}]$, where $L$ is the length of lookback windows and $H$ is the length of horizon windows. 
The univariate setting can be easily extended to the multivariate setting.
Let $ \{ {x}_t^{(1)}, {x}_t^{(2)}, \cdots, {x}_t^{(N)} \}_{t=1}^{T}$ stands for $N$ distinct time series with the same length $T$, and the {\textit{multivariate time series forecasting}} is: 
\begin{align}
    (\bm{x}^{(1)}_{t:t+H},\cdots, \bm{x}^{(N)}_{t:t+H})^T = \mathscr{F}_\Theta \left((\bm{x}^{(1)}_{t-L:t},\cdots, \bm{x}^{(N)}_{t-L:t})^T \right)
    \label{eq:base_ar}
\end{align}
where Gaussian noises $\bm{\epsilon}_{t:t+H}$ exist in the forecasting but dropped for brevity;
$\{ \bm{x}^{(i)}_{t-L:t} \}_{i=1}^{N}$ and $\{ \bm{x}^{(i)}_{t:t+H} \}_{i=1}^{N}$ are the multivariate lookback window and horizon window respectively; the mapping function $\mathscr{F}_{\Theta}: \R^{L \times N} \rightarrow \R^{H\times N}$ can be regarded as a forecasting model parameterized by $\Theta$.

\paragraph{Distribution Shift in Time Series.} As aforementioned, this paper focuses on two kinds of distribution shift in time series. In training forecasting models, one series will be cut into several lookback windows $\{ \bm{x}^{(i)}_{t-L:t} \}_{t=L}^{T-H}$ and their corresponding horizon windows $\{ \bm{x}^{(i)}_{t:t+H} \}_{t=L}^{T-H}$.
The \textbf{\textit{intra-space shift}} is defiend as: for any time-step $u \neq v$, 
\begin{equation}
    |d(\mathcal{X}^{(i)}_{input}(u) , \mathcal{X}^{(i)}_{input}(v))|>\delta
\end{equation}
where $\delta$ is a small threshold; $d$ is a distance function (\textit{e.g.,} KL divergence); $\mathcal{X}^{(i)}_{input}(u)$ and $ \mathcal{X}^{(i)}_{input}(v)$, standing for the distributions of lookback windows $\bm{x}^{(i)}_{u-L:u}$ and $\bm{x}^{(i)}_{v-L:v}$, are shifted.
Note that when most existing works \cite{ogasawara2010adaptive,wang2019deep, du2021adarnn,kim2022reversible} mention distribution shift in series, they mean our called intra-space shift. In contrast, the \textbf{\textit{inter-space shift}} is:
\begin{equation}
    |d(\mathcal{X}^{(i)}_{input}(u) , \mathcal{X}^{(i)}_{output}(u))|>\delta
\end{equation}
where $\mathcal{X}^{(i)}_{input}(u)$  and $\mathcal{X}^{(i)}_{output}(u)$ denotes the distribution of lookback window and horizon window at step $u$, respectively, which is ignored by current TSF models.

\section{\textsc{Dish-TS}}
In this section, we elaborate on our general neural paradigm, \textit{Dish-TS}. We start with an overview of this paradigm in Section \ref{sec:overview}. Then, we illustrate the architectures of \textit{Dish-TS} in Section \ref{sec:framework}. 
Also, we provide a simple and intuitive instance of \textit{Dish-TS} in Section \ref{sec:instance} and introduce a prior knowledge-induced training strategy in Section \ref{sec:training}, to demonstrate a workable design against the shift in forecasting.

\begin{figure}[th]
\centering
\includegraphics[width=0.97\linewidth]{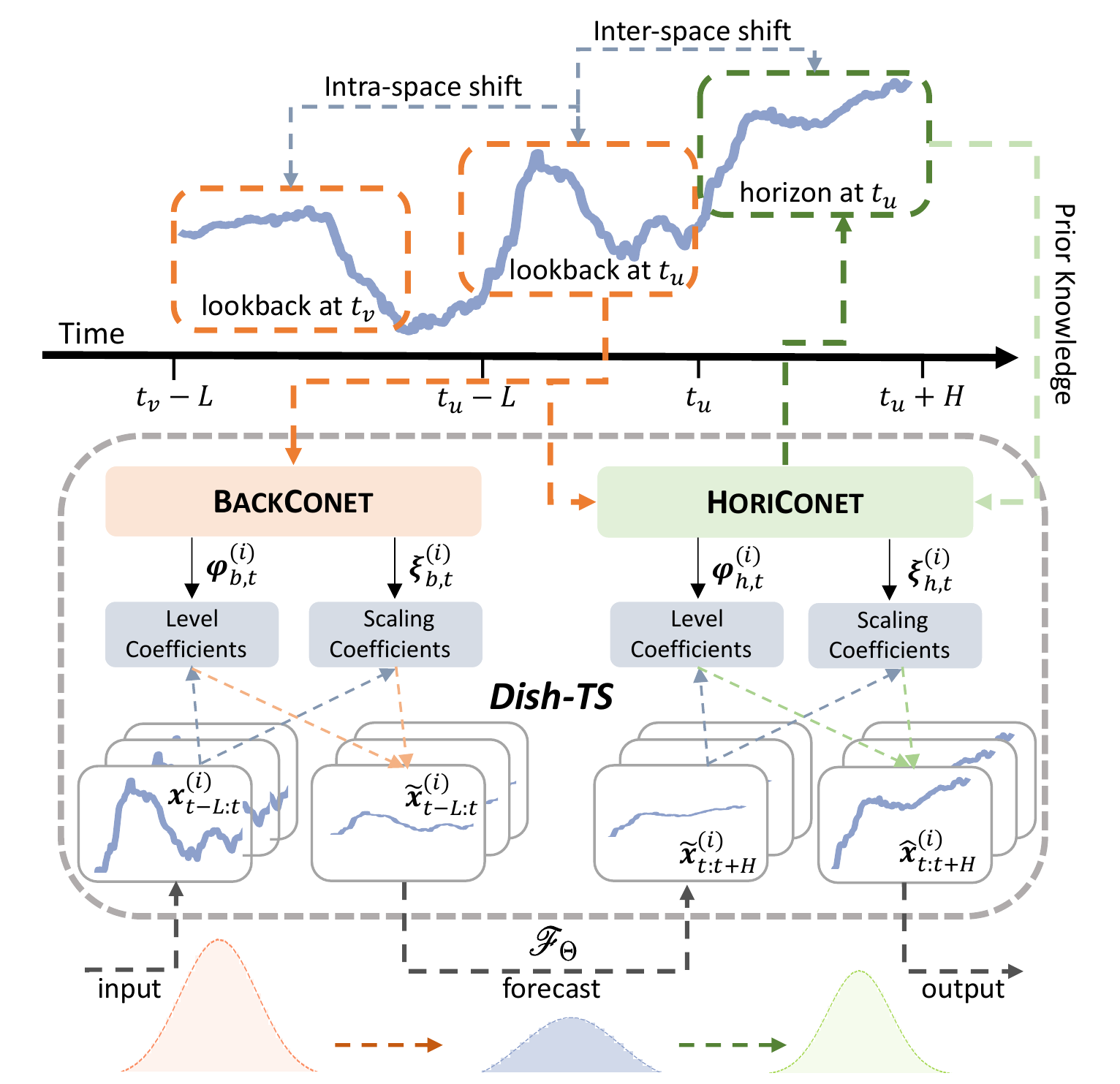}
\caption{Overview of Paradigm \textit{Dish-TS}.}
\label{fig:framework}
\end{figure}

\subsection{Overview} \label{sec:overview}
\textit{Dish-TS} is a simple yet effective, flexible paradigm against distribution shift in time series forecasting. Inspired by \cite{kim2022reversible}, \textit{Dish-TS} includes one two-stage process, normalizing before forecasting and denormalizing after forecasting.
The paradigm is built upon the coefficient net 
(\textsc{Conet}), 
which maps input series into coefficients for distribution measurement. 
As Figure \ref{fig:framework} shows, \textit{Dish-TS} is organized as a
dual-\textsc{Conet} framework, 
including a \textsc{BackConet} to illustrate input-space (lookbacks) and a \textsc{HoriConet} to illustrate output-space (horizons).
Data of lookbacks are transformed by coefficients from \textsc{BackConet} before being taken to any forecasting model $\mathscr{F}_\Theta$; the output ({\it i.e.}, forecasting results) are transformed by coefficients from \textsc{HoriConet} to acquire the final predictions. In addition, the \textsc{HoriConet} can be trained in a prior knowledge-induced fashion as a more effective way, especially in long series forecasting.

\subsection{Dual-{Conet} Framework}  \label{sec:framework}

We introduce \textsc{Conet} and Dual-\textsc{Conet} framework; then we illustrate how forecasting models are integrated into Dual-\textsc{Conet} by a two-stage normalize-denormalize process.


\subsubsection{{Conet}.}
Non-stationary time series makes it intractable for accurate predictions.
Pilot works \cite{ogasawara2010adaptive, du2021adarnn, kim2022reversible} measure distribution and its change via statistics (typically mean and std.) or distance function.
However, as stated in Section~\ref{sec:intro}, these operations are unreliable quantifications and limited in expressiveness. In this regard, we propose a coefficient net (\textsc{Conet}) for learning better distribution measurement to capture the shift. The general formulation is:
\begin{equation}
    \bm{\varphi}, \bm{\xi} = \textsc{Conet}( \bm{x} )
\end{equation}
where $\bm{\varphi} \in \R^1$ denotes \textit{level coefficient}, representing the overall scale of input series in a window $\bm{x} \in \R^L$; $\bm{\xi} \in \R^1$ denotes \textit{scaling coefficient}, representing fluctuation scale of $\bm{x}$.
In general, \textsc{Conet} could be set as any neural architectures  to conduct any linear/nonlinear mappings, which brings sufficient modeling capability and flexibility.

\subsubsection{Dual-{Conet}.} To relieve the aforementioned intra-space shift and inter-space shift in time series, \textit{Dish-TS} needs to capture the distribution difference among input-space and the difference between input-space and output-space. Inspired by one remarkable model N-BEATS \cite{Oreshkin2020N-BEATS} that uses `backcast' and `forecast' to conduct backward and forward predictions, we formulate the \textit{Dish-TS} as a Dual-\textsc{Conet} architecture, including 
a \textsc{BackConet} for input-space distribution of 
$ \{\bm{x}^{(i)}_{t-L:t}\}_{t=L}^{T-H} \in \mathcal{X}^{(i)}_{input}$, and
a \textsc{HoriConet} for output-space distribution of 
$ \{\bm{x}^{(i)}_{t:t+H} \}_{t=L}^{T-H} \in \mathcal{X}^{(i)}_{output}$.
In multivariate forecasting, the two \textsc{Conet}s are illustrated as:
\begin{equation}
\begin{aligned}
    \bm{\varphi}^{(i)}_{b,t}, \bm{\xi}^{(i)}_{b,t}  = \textsc{BackConet}( \bm{x}^{(i)}_{t-L:t} )&, i=1, \cdots, N  \\
    \bm{\varphi}^{(i)}_{h,t}, \bm{\xi}^{(i)}_{h,t} = \textsc{HoriConet}( \bm{x}^{(i)}_{t-L:t} )&, \; i=1, \cdots, N  
\end{aligned}
\end{equation}
where $\bm{\varphi}^{(i)}_{b,t}, \bm{\xi}^{(i)}_{b,t} \in \R^1$ are coefficients  for lookbacks, and $\bm{\varphi}^{(i)}_{h,t}, \bm{\xi}^{(i)}_{h,t} \in \R^1$ are coefficients for horizons at time-step $t$, given single $i$-th variate series.
Though sharing the same  input $\bm{x}^{(i)}_{t-L:t}$, the two \textsc{Conet}s have distinct targets, where \textsc{BackConet} aims to approximate distribution $\mathcal{X}^{(i)}_{input}$ from input lookback signals, while \textsc{HoriConet} is to infer (or predict) future distribution $\mathcal{X}^{(i)}_{output}$ based on historical observations. This brings additional challenges in training \textsc{HoriConet}, detailed in Section \ref{sec:training}.

\subsubsection{Integrating Dual-{Conet} into Forecasting.}
After acquiring coefficients from Dual-\textsc{Conet}, the coefficients can be integrated into any time series forecasting model to alleviate the two aforementioned shifts through a two-stage normalizing-denormalizing process.
Specifically, let $\mathscr{F}_\Theta$ represent any forecasting model, the original forecasting process $\hat{\bm{x}}^{(i)}_{t:t+H} =  \mathscr{F}_\Theta(\bm{x}^{(i)}_{t-L:t})$ is rewritten as:
\begin{equation} \label{eq:normdenorm}
\hat{\bm{x}}^{(i)}_{t:t+H} =  \bm{\xi}^{(i)}_{h,t} \mathscr{F}_\Theta \left( \frac{1}{\bm{\xi}^{(i)}_{b,t}} (\bm{x}^{(i)}_{t-L:t} - \bm{\varphi}^{(i)}_{b,t}) \right) + \bm{\varphi}^{(i)}_{h,t}
\end{equation}
where $\hat{\bm{x}}^{(i)}_{t:t+H}$ are the final transformed forecasting results after integration with dual conets. Actually, Equation (\ref{eq:normdenorm}) includes a two-stage process with $\mathscr{F}_\Theta$: (i) normalize input lookbacks $\bm{x}^{(i)}_{t-L:t}$ before forecasting  by $ \tilde{\bm{x}}^{(i)}_{t-L:t} = \frac{1}{\bm{\xi}^{(i)}_{b,t}} (\bm{x}^{(i)}_{t-L:t} - \bm{\varphi}^{(i)}_{b,t})$;
(ii) denormalize model's direct output $\tilde{{\bm{x}}}^{(i)}_{t:t+H}$ after forecasting by $\hat{\bm{x}}^{(i)}_{t:t+H} = \bm{\xi}^{(i)}_{h,t}\tilde{\bm{x}}^{(i)}_{t:t+H} + \bm{\varphi}^{(i)}_{h,t} $. Note that even though the above operations only consider additive and multiplicative transformations, \textsc{Conet} itself could be any complicated linear/nonlinear mappings in the generation of coefficients, which is flexible. Finally, the transformed forecasts 
$\hat{\bm{x}}^{(i)}_{t:t+H}$ are taken to loss optimization.

\subsection{A Simple and Intuitive Instance of {Conet}} \label{sec:instance}

Essentially, the flexibility of \textit{Dish-TS} comes from the specific \textsc{Conet} design, which could be any neural architectures for different modeling capacity. To demonstrate the effectiveness of our framework, we provide a most simple and intuitive instance of \textsc{Conet} design to reduce series shift.

Specifically, given multivariate input $\{\bm{x}^{(i)}_{t-L:t}\}_{i=1}^{N}$, the most intuitive way is to use standard fully connected layers to conduct linear projections. 
Let $\mathbf{v}^{\ell}_b,\mathbf{v}^{\ell}_h  \in \R^{L*N}$ stand for two basic learnable vectors of layer $\ell$ of \textsc{BackConet} and \textsc{HoriConet} respectively. Here we consider $\ell=1$ for simplicity, and then the projection is:
\begin{equation}
   \bm{\varphi}^{(i)}_{b,t} =  \sigma ( \sum_{\tau=1}^{\text{dim}(\mathbf{v}^{\ell}_{b, i})} \mathbf{v}^{\ell}_{b, i\tau}   {x}^{(i)}_{\tau-L+t}),\\
   \bm{\varphi}^{(i)}_{h,t} =  \sigma ( \sum_{\tau=1}^{\text{dim}(\mathbf{v}^{\ell}_{h, i})} \mathbf{v}^{\ell}_{h, i\tau}   {x}^{(i)}_{\tau-L+t})
\end{equation}
where the level coefficients $\bm{\varphi}^{(i)}_{b,t}$ and  $\bm{\varphi}^{(i)}_{h,t}$ are respectively from \textsc{BackConet} and \textsc{HoriConet} to represent the overall scale of input $\bm{x}^{(i)}_{t-L:t}$ and output $\bm{x}^{(i)}_{t:t+H}$; $\sigma$ here denotes a leaky ReLU non-linearity \cite{maas2013rectifier} is utilized instead of original ReLU that ignores negative data and thus causes information loss.
Also, we aim to let scaling coefficients represent the fluctuation for series. Inspired by the calculation of standard deviation $\sqrt{\int_{-\infty}^{+\infty}(x-\mu)^{2} f(x) d x}$ where $x$ is variable and $\mu$ is mean, we propose the following operation to get scaling coefficients:
\begin{equation} \label{eq:var}
 \bm{\xi}^{(i)}_{b,t} =    \sqrt{ \mathbb{E} ( {x}^{(i)}_t - \bm{\varphi}^{(i)}_{b,t}  )^2 }, \;\;
 \bm{\xi}^{(i)}_{h,t} =    \sqrt{ \mathbb{E}  ( {x}^{(i)}_t - \bm{\varphi}^{(i)}_{h,t}  )^2 }
\end{equation}
where scaling coefficients $\bm{\xi}^{(i)}_{b,t}, \bm{\xi}^{(i)}_{h,t}$ can actually be seen as the average deviation of $\bm{x}^{(i)}_{t-L:t}$ with regard to $\bm{\varphi}^{(i)}_{b,t}$ and $\bm{\varphi}^{(i)}_{h,t}$.
The equation (\ref{eq:var}) is also simple, intuitive and easy to compute, without introducing extra parameters to optimize.

\subsection{Prior Knowledge-Induced Training Strategy}
\label{sec:training}

As aforementioned, 
\textsc{BackConet} estimates distribution of input-space $\mathcal{X}^{(i)}_{input}$, while
\textsc{HoriConet} needs to infer distribution of output-space $\mathcal{X}^{(i)}_{output}$, which is more intractable because of the gap between input- and output-space. The gap is even larger with the increase of horizon length. 

To solve this problem, we aim to pour some prior knowledge ({\it i.e.,} mean of horizons) as soft targets in \textit{Dish-TS} to assist the learning of \textsc{HoriConet} to generate coefficients $\bm{\varphi}^{(i)}_{h,t},\bm{\xi}^{(i)}_{h,t}$.
Even though the statistic mean of horizons cannot fully reflect the distribution, it can still demonstrate characteristics of output-space, as discussed in Section \ref{sec:intro}. 
Thus, along the line of equation (\ref{eq:normdenorm}), the classic mean square error can be given by
$\mathcal{L}_{mse} = \sum_{k=1}^{K} \sum_{i=1}^{N} \left( \hat{\bm{x}}^{(i)}_{t_k:t_k+H} - {\bm{x}}^{(i)}_{t_k:t_k+H}   \right)^2$, where $K$ is the batch size, $t_k$ is randomly-sampled time points to compose batches, and $N$ is number of series. With prior knowledge, we rewrite the final optimization loss $\mathcal{L}$ as:
\begin{equation} \label{eq:distillation}
  \sum_{k=1}^{K} \sum_{i=1}^{N}  [
 { (\hat{\bm{x}}^{(i)}_{t_k:t_k+H}-{\bm{x}}^{(i)}_{t_k:t_k+H} )^2
 }
  + \underbrace{ 
 \alpha (\frac{1}{H} \sum_{t=t_k+1}^{t_k+H} x^{(i)}_{t} - \bm{\varphi}^{(i)}_{h,t_k})^2 
 }_{\text{Prior Knowledge Guidance}}  
]
\end{equation}
where the left item is mean square error; the right item is the learning guidance of prior knowledge; $\alpha$ is to control weight of prior guidance; $\bm{\varphi}^{(i)}_{h,t_k}$ is the level coefficients of \textsc{HoriConet} to softly optimize.

\begin{table*}[h] 
\centering
\footnotesize
\begin{threeparttable}
\small
\fontsize{8.9pt}{9.7}\selectfont
\caption{Univariate time series forecasting performance. The length of lookbacks/horizons is set the same.} \label{table:univariatev2}
\begin{tabular}{c|c|cccc|cccc|cccc}
\toprule
\multicolumn{2}{c}{Method}& \multicolumn{2}{c}{Informer}&\multicolumn{2}{c}{+\textit{Dish-TS}}& \multicolumn{2}{c}{Autoformer}&\multicolumn{2}{c}{+\textit{Dish-TS}}& 
\multicolumn{2}{c}{N-BEATS}&   \multicolumn{2}{c}{+\textit{Dish-TS}}\\ 
\cmidrule(r){1-2} \cmidrule(r){3-4} \cmidrule(r){5-6} \cmidrule(r){7-8} \cmidrule(r){9-10} \cmidrule(r){11-12} \cmidrule(r){13-14} 
\multicolumn{2}{c}{Metric}&
MSE& MAE&  MSE& MAE&
MSE& MAE&  MSE& MAE&
MSE& MAE&  MSE& MAE\\
\midrule \midrule
    \multirow{5}{*}{ \rotatebox{90}{Electricity} }
&24&4.394 & 4.897 & \textbf{1.116} & \textbf{2.413}&1.616 & 3.138 & \textbf{1.535} & \textbf{2.942}&1.259 & 2.572 & \textbf{1.205} & \textbf{2.536}\\
&48&4.405 & 4.904 & \textbf{1.256} & \textbf{2.581}&2.291 & 3.772 & \textbf{1.783} & \textbf{3.199}&1.311 & 2.696 & \textbf{1.286} & \textbf{2.629}\\
&96&3.933 & 4.675 & \textbf{1.060} & \textbf{2.354}&2.281 & 3.726 & \textbf{1.330} & \textbf{2.706}&1.271 & 2.625 & \textbf{1.129} & \textbf{2.447}\\
&168&4.083 & 4.747 & \textbf{1.015} & \textbf{2.301}&2.072 & 3.502 & \textbf{1.162} & \textbf{2.413}&1.520 & 2.879 & \textbf{0.985} & \textbf{2.261}\\
&336&4.292 & 4.848 & \textbf{2.893} & \textbf{4.364}&2.112 & 3.481 & \textbf{1.387} & \textbf{2.722}&1.747 & 3.107 & \textbf{1.497} & \textbf{2.852}\\
\midrule
\multirow{5}{*}{ \rotatebox{90}{ETTh1} }
&24&0.427 & 1.536 & \textbf{0.339} & \textbf{1.343}&0.420 & 1.505 & \textbf{0.344} & \textbf{1.350}&0.406 & 1.479 & \textbf{0.320} & \textbf{1.279}\\
&48&0.855 & 2.272 & \textbf{0.570} & \textbf{1.824}&0.767 & 2.172 & \textbf{0.588} & \textbf{1.885}&0.598 & 1.841 & \textbf{0.569} & \textbf{1.815}\\
&96&0.930 & 2.328 & \textbf{0.840} & \textbf{2.258}&1.100 & 2.669 & \textbf{0.872} & \textbf{2.340}&0.827 & 2.254 & \textbf{0.785} & \textbf{2.168}\\
&168&0.964 & 2.567 & \textbf{0.911} & \textbf{2.448}&1.098 & 2.659 & \textbf{0.959} & \textbf{2.488}&1.052 & 2.617 & \textbf{0.908} & \textbf{2.399}\\
&336&1.146 & 2.829 & \textbf{0.993} & \textbf{2.520}&1.230 & 2.796 & \textbf{0.936} & \textbf{2.454}&1.117 & 2.639 & \textbf{1.011} & \textbf{2.550}\\
\midrule
\multirow{5}{*}{ \rotatebox{90}{ETTm2} }
&24&1.328 & 2.525 & \textbf{0.760} & \textbf{1.686}&1.718 & 2.976 & \textbf{0.762} & \textbf{1.851}&0.867 & 1.804 & \textbf{0.600} & \textbf{1.479}\\
&48&1.488 & 2.649 & \textbf{1.070} & \textbf{2.117}&3.061 & 4.259 & \textbf{1.847} & \textbf{3.082}&1.290 & 2.374 & \textbf{1.145} & \textbf{2.160}\\
&96&2.952 & 4.324 & \textbf{1.631} & \textbf{2.771}&3.113 & 4.309 & \textbf{2.385} & \textbf{3.648}&1.707 & 2.922 & \textbf{1.605} & \textbf{2.747}\\
&168&5.114 & 5.832 & \textbf{2.754} & \textbf{3.841}&4.167 & 4.959 & \textbf{3.413} & \textbf{4.452}&2.428 & 3.603 & \textbf{2.380} & \textbf{3.579}\\
&336&5.958 & 6.490 & \textbf{4.284} & \textbf{5.096}&5.753 & 5.993 & \textbf{4.449} & \textbf{5.213}&3.974 & 4.815 & \textbf{3.568} & \textbf{4.582}\\
\midrule
\multirow{5}{*}{ \rotatebox{90}{Weather} }
&24&3.632 & 1.381 & \textbf{0.725} & \textbf{0.584}&1.082 & 0.775 & \textbf{0.800} & \textbf{0.622}&0.570 & 0.486 & \textbf{0.567} & \textbf{0.480}\\
&48&5.933 & 1.856 & \textbf{1.251} & \textbf{0.798}&1.617 & 0.968 & \textbf{1.317} & \textbf{0.845}&1.272 & 0.825 & \textbf{1.178} & \textbf{0.776}\\
&96&6.895 & 2.071 & \textbf{1.898} & \textbf{1.022}&1.901 & 1.034 & \textbf{1.824} & \textbf{1.005}&1.898 & 0.995 & \textbf{1.783} & \textbf{0.994}\\
&168&6.786 & 2.045 & \textbf{1.932} & \textbf{1.042}&1.970 & 1.046 & \textbf{1.847} & \textbf{1.008}&2.571 & 1.210 & \textbf{1.848} & \textbf{1.024}\\
&336&7.393 & 2.175 & \textbf{2.237} & \textbf{1.099}&2.190 & 1.100 & \textbf{2.015} & \textbf{1.061}&3.624 & 1.486 & \textbf{2.447} & \textbf{1.117}\\
\midrule
\bottomrule
\end{tabular}
\begin{tablenotes}
\item Results of \textit{Illness} dataset are included Appendix \ref{sec:app_B.1}, due to space limit.
\end{tablenotes}
\end{threeparttable}
\end{table*}

\section{Experiment}

\subsection{Experimental Setup}

\subsubsection{Datasets.} 
We conduct our experiments on five real-world datasets: 
(i) \textbf{\textit{Electricity}} 
dataset collects the electricity consumption (Kwh) of 321 clients. 
(ii) {\textit{ETT}} 
 dataset includes data of electricity transformers temperatures. We select \textbf{\textit{ETTh1}} dataset (hourly) and \textbf{\textit{ETTm2}} dataset (15-minutely). 
(iii) \textbf{\textit{Weather}} 
 dataset records 21 meteorological features every ten minutes. 
(iv) \textbf{\textit{Illness}} 
dataset includes weekly-recorded influenza-like illness patients data. We mainly follow \cite{zhou2021informer} and \cite{xu2021autoformer} to preprocess and split data. More details are in Appendix \ref{sec:app_A.1}

\subsubsection{Evaluation.} 
To directly reflect distribution shift in time series, all the experiments are conducted on original data without data normalization or scaling.
We evaluate time series forecasting performance on the mean squared error (MSE) and mean absolute error (MAE). Note that our evaluations are on original data; thus the reported metrics are scaled for readability. More evaluation details are in Appendix \ref{sec:app_A.2}.

\subsubsection{Implementation.}
All the experiments are implemented with PyTorch \cite{paszke2019pytorch} on an NVIDIA RTX 3090 24GB GPU. 
In training, all the models are trained using L2 loss and Adam \cite{kingma2014adam} optimizer with learning rate of [1e-4, 1e-3]. We repeat three times for each experiment and report average performance. 
We let lookback/horizon windows have the same length, gradually prolonged from 24 to 336 except for \textit{illness} dataset that has length limitation. We have also discussed larger lookback length $L$, larger horizon length $H$, and prior guidance rate $\alpha$. Implementation details are included in Appendix \ref{sec:app_A.3}.

\subsubsection{Baselines.} As aforementioned, our \textit{Dish-TS} is a general neural framework that can be integrated into any deep time series forecasting models for end-to-end training. To verify the effectiveness, we couple our paradigm with three state-of-the-art backbone models, Informer \cite{zhou2021informer}, Autoformer \cite{xu2021autoformer} and N-BEATS \cite{Oreshkin2020N-BEATS}. 
More baseline details are in Appendix \ref{sec:app_A.4}.

\begin{table*}[ht]
\centering
\begin{threeparttable}
\fontsize{8.9pt}{9.7}\selectfont
\caption{Multivariate time series forecasting performance of backbones and \textit{Dish-TS}.
} \label{table:multivariatev2}
\begin{tabular}{c|c|cccc|cccc|cccc}
\toprule
\multicolumn{2}{c}{Method}& \multicolumn{2}{c}{Informer}&\multicolumn{2}{c}{+\textit{Dish-TS}}& \multicolumn{2}{c}{Autoformer}&\multicolumn{2}{c}{+\textit{Dish-TS}}& 
\multicolumn{2}{c}{N-BEATS}&   \multicolumn{2}{c}{+\textit{Dish-TS}}\\ 
\cmidrule(r){1-2} \cmidrule(r){3-4} \cmidrule(r){5-6} \cmidrule(r){7-8} \cmidrule(r){9-10} \cmidrule(r){11-12} \cmidrule(r){13-14} 
\multicolumn{2}{c}{Metric}&
MSE& MAE&  MSE& MAE&
MSE& MAE&  MSE& MAE&
MSE& MAE&  MSE& MAE\\
\midrule \midrule
\multirow{5}{*}{ \rotatebox{90}{Electricity} }
&24&0.482 & 0.575 & \textbf{0.036} & \textbf{0.249}&0.082 & 0.420 & \textbf{0.040} & \textbf{0.247}&0.041 & 0.281 & \textbf{0.032} & \textbf{0.241}\\
&48&0.969 & 1.002 & \textbf{0.056} & \textbf{0.289}&0.125 & 0.450 & \textbf{0.051} & \textbf{0.278}&0.043 & 0.275 & \textbf{0.041} & \textbf{0.265}\\
&96&1.070 & 1.046 & \textbf{0.084} & \textbf{0.325}&0.363 & 0.642 & \textbf{0.064} & \textbf{0.285}&0.067 & 0.324 & \textbf{0.058} & \textbf{0.286}\\
&168&0.960 & 1.013 & \textbf{0.088} & \textbf{0.335}&0.585 & 0.835 & \textbf{0.080} & \textbf{0.319}&0.078 & 0.347 & \textbf{0.074} & \textbf{0.294}\\
&336&1.113 & 1.058 & \textbf{0.153} & \textbf{0.400}&0.569 & 0.766 & \textbf{0.104} & \textbf{0.357}&0.108 & 0.383 & {0.108} & \textbf{0.355}\\
\midrule
\multirow{5}{*}{ \rotatebox{90}{ETTh1} }
&24&0.988 & 1.794 & \textbf{0.876} & \textbf{1.633}&1.451 & 2.100 & \textbf{1.019} & \textbf{1.744}&0.797 & 1.531 & \textbf{0.790} & \textbf{1.501}\\
&48&1.318 & 2.127 & \textbf{1.073} & \textbf{1.846}&1.456 & 2.161 & \textbf{1.240} & \textbf{1.966}&0.913 & 1.696 & \textbf{0.907} & \textbf{1.657}\\
&96&2.333 & 2.965 & \textbf{1.185} & \textbf{2.011}&1.371 & 2.173 & \textbf{1.199} & \textbf{1.982}&1.057 & 1.875 & \textbf{0.975} & \textbf{1.793}\\
&168&2.778 & 3.234 & \textbf{1.273} & \textbf{2.085}&1.267 & 2.146 & \textbf{1.148} & \textbf{1.991}&1.038 & 1.893 & \textbf{0.994} & \textbf{1.858}\\
&336&2.825 & 3.335 & \textbf{1.779} & \textbf{2.586}&1.334 & 2.333 & \textbf{1.147} & \textbf{2.062}&1.128 & 2.020 & \textbf{1.055} & \textbf{1.976}\\
\midrule
\multirow{5}{*}{ \rotatebox{90}{ETTm2} }
&24&1.352 & 2.443 & \textbf{0.608} & \textbf{1.594}&0.834 & 1.882 & \textbf{0.676} & \textbf{1.701}&0.643 & 1.609 & \textbf{0.634} & \textbf{1.587}\\
&48&1.781 & 2.973 & \textbf{0.736} & \textbf{1.767}&1.165 & 2.269 & \textbf{0.823} & \textbf{1.903}&0.808 & 1.829 & \textbf{0.785} & \textbf{1.793}\\
&96&1.936 & 3.017 & \textbf{0.877} & \textbf{1.946}&1.165 & 2.237 & \textbf{0.929} & \textbf{2.021}&0.953 & 2.017 & \textbf{0.860} & \textbf{1.904}\\
&168&2.822 & 3.656 & \textbf{1.213} & \textbf{2.273}&1.404 & 2.423 & \textbf{1.308} & \textbf{2.367}&1.094 & 2.143 & \textbf{1.087} & \textbf{2.156}\\
&336&2.778 & 3.638 & \textbf{1.620} & \textbf{2.637}&1.795 & 2.739 & \textbf{1.603} & \textbf{2.624}&1.498 & 2.543 & \textbf{1.448} & \textbf{2.522}\\
\midrule
\multirow{5}{*}{ \rotatebox{90}{Weather} }
&24&3.552 & 2.120 & \textbf{2.224} & \textbf{1.100}&4.485 & 2.313 & \textbf{2.481} & \textbf{1.163}&2.557 & 1.454 & \textbf{2.267} & \textbf{1.093}\\
&48&4.206 & 2.231 & \textbf{3.610} & \textbf{1.644}&6.581 & 2.815 & \textbf{4.299} & \textbf{1.781}&5.527 & 2.393 & \textbf{4.783} & \textbf{1.802}\\
&96&3.064 & 2.042 & \textbf{2.507} & \textbf{1.457}&5.812 & 2.569 & \textbf{3.280} & \textbf{1.806}&2.539 & 1.566 & \textbf{2.280} & \textbf{1.367}\\
&168&2.713 & 2.106 & \textbf{2.184} & \textbf{1.450}&4.053 & 2.188 & \textbf{3.309} & \textbf{1.950}&2.160 & 1.604 & \textbf{1.885} & \textbf{1.309}\\
&336&3.472 & 2.567 & \textbf{2.238} & \textbf{1.611} &3.910 & 2.111 & \textbf{3.314} & \textbf{1.982}& 2.043 & 1.551 & \textbf{1.757} & \textbf{1.366}\\
\midrule
\bottomrule
\end{tabular}
\begin{tablenotes}
\item $*$ means N-BEATS is re-implemented for multivariate time series forecasting; see Appendix \ref{sec:app_A.4} for more details.
\end{tablenotes}
\end{threeparttable}
\end{table*}

\begin{table*}[h]
\small
\centering
\caption{Performance comparisons on MSE with the state-of-the-art normalization technique in multivariate time series forecasting taking Autoformer as the backbone. Univariate forecasting results are in Appendix \ref{sec:app_B.2}.}
\begin{tabular}{c|ccc|ccc|ccc|ccc}
\toprule
Datasets & \multicolumn{3}{c}{Electiricity} & \multicolumn{3}{c}{ETTh1} & \multicolumn{3}{c}{ETTm2} & \multicolumn{3}{c}{Weather}  \\
Length & 24 & 168 & 336 & 24 & 168 & 336 & 24 & 168 & 336 & 24 & 168 & 336  \\
\midrule
RevIN & 0.044 & 0.091 & 0.109 & 1.245 & 1.462 & 1.920 & 0.794 & 1.501 & 1.827 &
3.523 & 3.658 & 3.501\\
\textit{Dish-TS}&0.039 & 0.076 & 0.086 & 1.018 & 1.148 & 1.222 & 0.651 & 1.325 & 1.599 &
2.481 & 3.283 & 3.232\\
\midrule
Improve & 11.3\%& 15.3\%& 21.1\%& 18.1\%& 21.4\%& 36.3\%& 
17.7\%& 11.7\%& 12.3\%&
29.6\%&  10.7\%&  7.7\% \\
\bottomrule
\end{tabular}
\label{table:norm_compare_M}
\end{table*}

\subsection{Overall Performance}
\textbf{\textit{Univariate time series forecasting}}. Table \ref{table:univariatev2} demonstrates the overall univariate time series forecasting performance of three state-of-the-art backbones and their \textit{Dish-TS} equipped versions, where we can easily observe that \textit{Dish-TS} helps all the backbones achieve much better performance. The most right column of Table \ref{table:univariatev2} shows the average improvement of \textit{Dish-TS} over baseline models under different circumstances. We can see that \textit{Dish-TS} can achieve a MSE improvement more than 20\% in most cases, up to 50\% in some cases. Notably, Informer usually performs worse but can be improved significantly with \textit{Dish-TS}.  

\noindent \textbf{\textit{Multivariate time series forecasting.}} Table \ref{table:multivariatev2} demonstrates the overall multivariate time series forecasting performance across four datasets and the results and analysis for \textit{Illness} dataset are in Appendix \ref{sec:app_B.1}. Still, we notice \textit{Dish-TS} can also improve significantly in the task of multivariate forecasting compared with three backbones. We find out a stable improvement (from 10\% to 30\%) on \textit{ETTh1}, \textit{ETTm2} and \textit{Weather} datasets when coupled with \textit{Dish-TS}.
Interestingly, we notice the both original Informer and Autoformer can hardly converge well in \textit{Electricity} original data. With \textit{Dish-TS}, the data distribution is normalized for better forecasting.

\subsection{Comparison with Normalization Methods}
In this section, we further compare performance with the state-of-the-art normalization technique, RevIN \cite{kim2022reversible},  that handles distribution shift in time series forecasting. Here we don't consider AdaRNN \cite{du2021adarnn} because it is not compatible for fair comparisons. 
Table \ref{table:norm_compare_M} shows the comparison results in multivariate time series forecasting. We can easily observe though RevIN actually improves performance of vanilla backbone (Autoformer) to some degree, \textit{Dish-TS} can still achieve more than 10\% improvement on average compared with RevIN. 
A potential reason for this significant improvement of such a simple \textsc{Conet} design is the consideration towards both intra-space shift and inter-space shift.

\begin{table}[h]
\centering
\fontsize{8.2pt}{9.7}\selectfont
\caption{Impact of larger horizons on forecasting, also referred as long TSF problems. Performance (MSE) is reported when horizon is prolonged (336 to 720) and lookback is fixed as 96, taking N-BEATS as backbone model.}
\label{table:horizon}
\begin{tabular}{c|cccccc}
\toprule 
Horizon&336&420&540&600&720 \\
\midrule
\textbf{Electricity}&1.7429&1.7859&1.7720&1.6140&1.6023  \\
+\textit{Dish-TS}&1.3361& 1.4507&1.4107&1.4340&1.4785\\
\midrule 
\textbf{ETTh1}& 1.0468&1.2688&1.1696&1.3281&1.4270 \\
+\textit{Dish-TS}&0.9699&1.0864&1.1361&1.1852&1.1913  \\
\bottomrule
\end{tabular}
\end{table}

\begin{figure*}[th]
\centering
\includegraphics[width=0.21\linewidth]{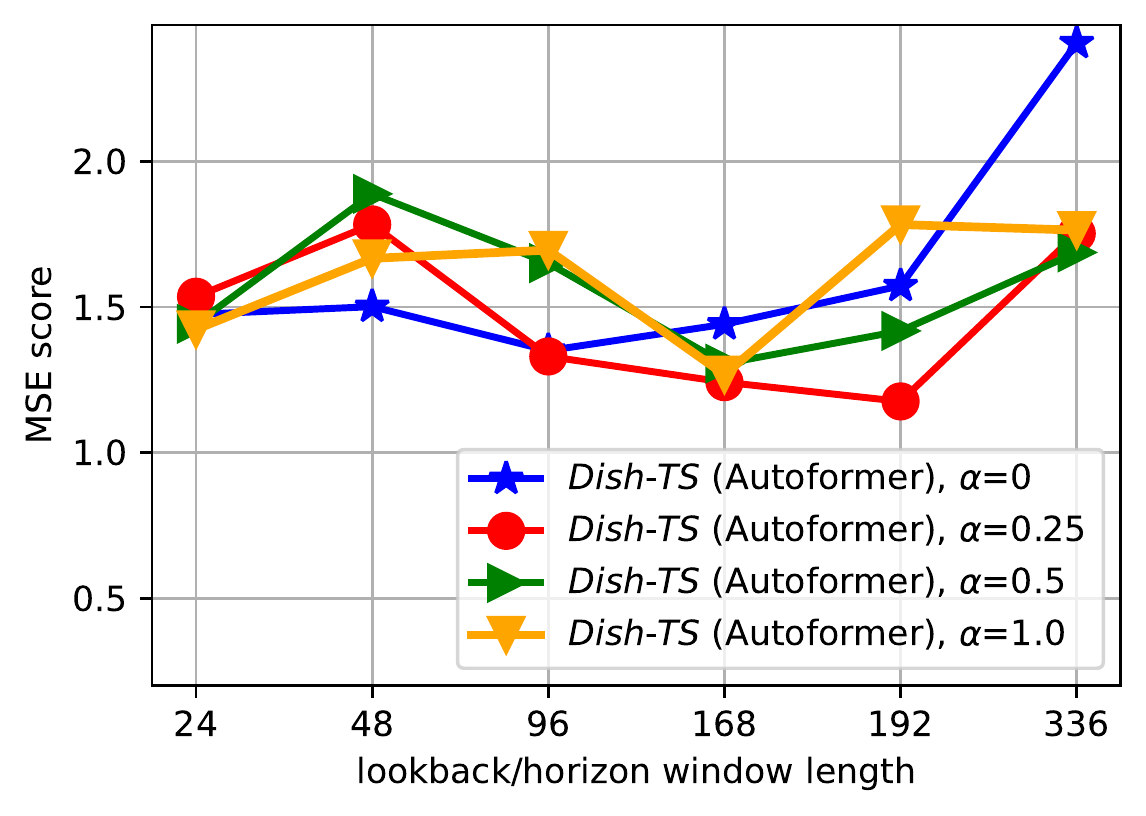}
\includegraphics[width=0.21\linewidth]{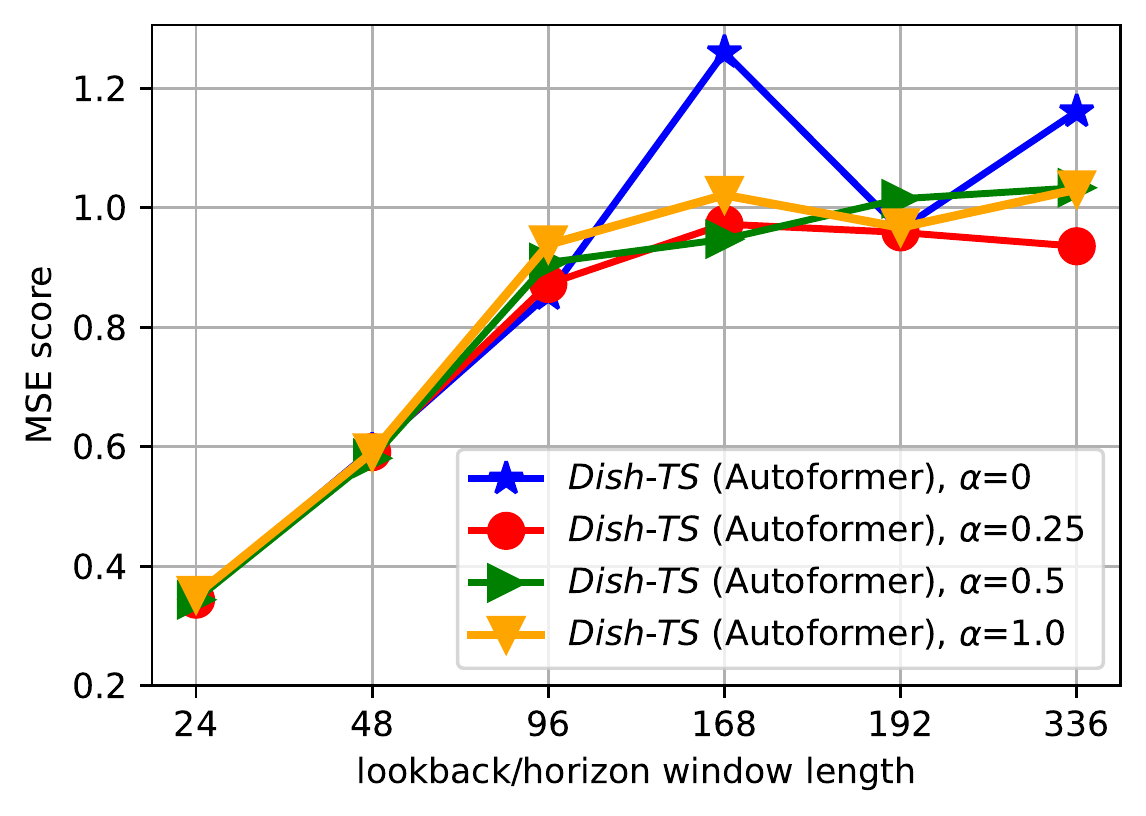}
\includegraphics[width=0.20\linewidth]{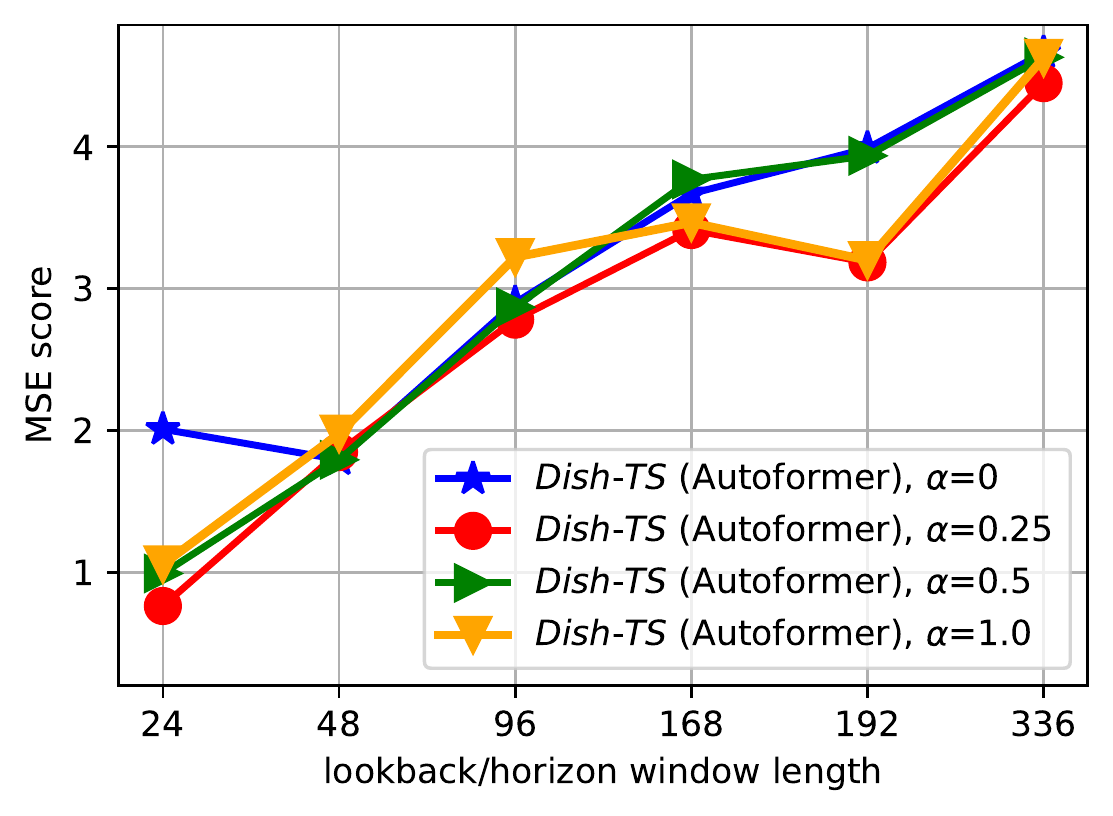}
\includegraphics[width=0.21\linewidth]{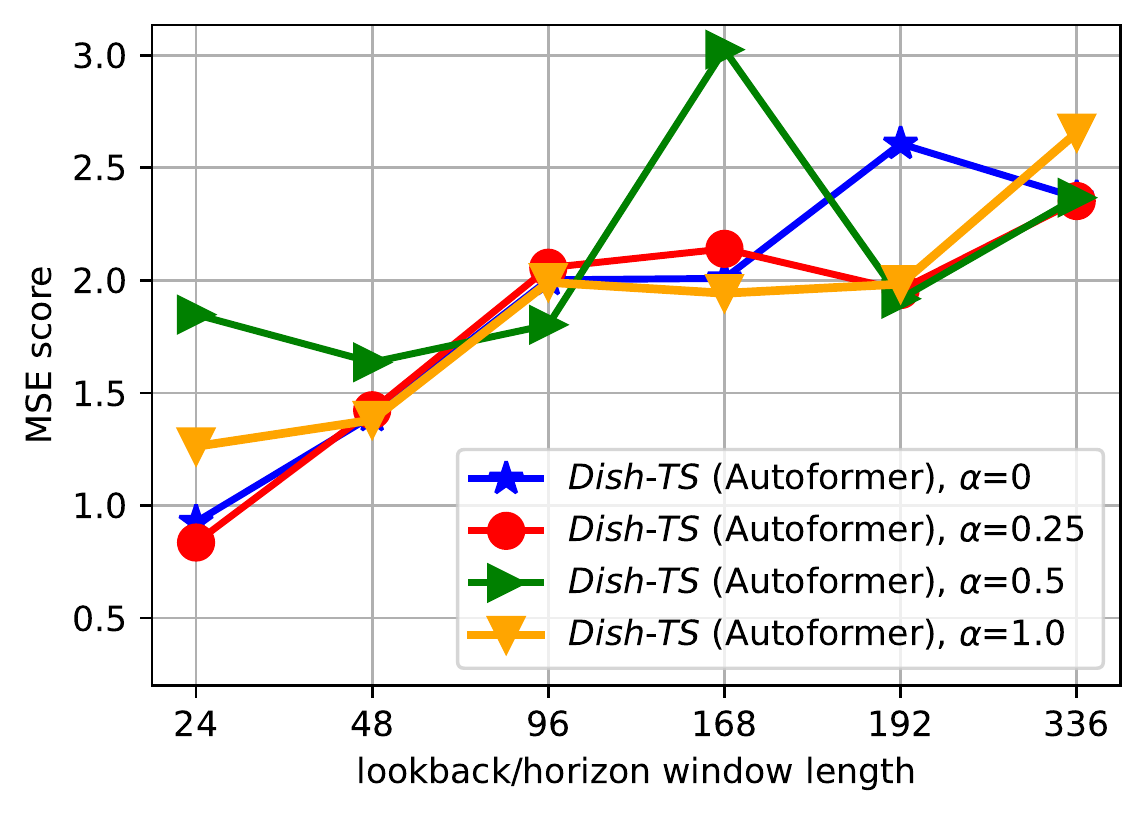}
\caption{Impact of prior guidance $\alpha$ on model performance on (from left to right) \textit{Electricity}, \textit{ETTh1}, \textit{ETTm2}, \textit{Weather} datasets.}
\label{figure:rate}
\end{figure*}

\subsection{Parameters and Model Analysis} \label{sec:parameter_model}

\subsubsection{Horizon Analysis.} 
We aim to discuss the influence of larger horizons (known as long time series forecasting \cite{zhou2021informer}) on the model performance.
Interestingly, from Table \ref{table:horizon}, we find out backbone (N-BEATS) performs even better in \textit{Electricity} as horizon becomes larger while on other datasets like \textit{ETTh1} larger horizons introduce more difficulty in forecasting. However, \textit{Dish-TS} can still achieve better performance in different settings. Performance on \textit{Dish-TS} is slowly worse with horizon's increasing. An intuitive reason is larger horizons include more distribution changes and thus need more complicated modeling.

\begin{table}[h]
\centering
\fontsize{8.2pt}{9.7}\selectfont
\caption{Impact of lookback length on forecasting. Metric MSE is reported when lookback is set from 48 to 240 when horizon is fixed as 48, taking N-BEATS as the backbone.}
\label{table:large_lookback}
\begin{tabular}{c|cccccc}
\toprule 
Lookback&48&96&144&192&240 \\
\midrule
\textbf{Electricity}& 1.3026 & 1.3673& 1.2794& 1.2686&0.9494\\
+\textit{Dish-TS}& 1.2862& 0.9682& 0.7309& 0.7156& 0.7605\\
\midrule 
\textbf{ETTh1}& 0.5979&0.5745&0.5459&0.5638&0.6309 \\
+\textit{Dish-TS}&0.5708&0.5451&0.5202&0.5307&0.5234  \\
\bottomrule
\end{tabular}
\end{table}

\subsubsection{Lookback Analysis.}
We analyze the influence of lookback length on the model performance. As Table \ref{table:large_lookback} shows, we notice \textit{Dish-TS} achieves $1.286 \rightarrow 0.731$ on \textit{Electricity} and $0.571 \rightarrow 0.520$ when lookback increases from 48 to 144. This signifies in many cases larger lookback windows  bring more historical information to infer the future distribution, thus boosting the prediction performance.

\subsubsection{Prior Guidance Rate.} We study the impact of prior guidance on model performance. Figure \ref{figure:rate} shows the performance comparison with different guidance weight $\alpha$ in Equation (\ref{eq:distillation}). 
From the table, we observe when lookback/horizon is small, the performance gap among different $\alpha$s is less obvious. However, when length is larger (than 168), the prediction error of $\alpha = 0$ (no guidance) increases quickly, while other settings achieve less errors. This shows prior knowledge can help better \textsc{HoriConet} learning, especially in the settings of long time series forecasting.

\begin{table}[h]
\centering
\fontsize{7.5pt}{9}\selectfont
\caption{Impact of initialization of \textsc{Conet}. Lookbacks and horizons have the same length. Underlined are best results.}
\label{table:init}
\begin{tabular}{c|c|ccc|ccc}
\toprule 
\multicolumn{2}{c}{Model}& \multicolumn{3}{c}{ \textit{Dish-TS} (Autoformer)}    &\multicolumn{3}{c}{\textit{Dish-TS} (N-BEATS)}    \\
\multicolumn{2}{c}{Initilize}& avg & norm & uniform & avg & norm & uniform \\
\midrule
\multirow{3}{*}{\rotatebox{90}{ETTh1} }
&24&\underline{3.439}&	3.658&	3.532&  \underline{3.196}&	3.230&	3.292\\
&96&8.794&	9.381&	\underline{8.774}& 7.878&	8.067&	\underline{7.851}\\
&168&9.878&	9.725&	\underline{9.589}& 9.783&	9.250&	\underline{9.080}\\
\midrule
\multirow{3}{*}{\rotatebox{90}{Weather} }
&24&1.579&	0.835&	\underline{0.799}& 0.650&	0.579&	\underline{0.566}\\
&96&2.127&	2.056&	\underline{1.823}& 1.915&	1.814&	\underline{1.782}\\
&168&2.139&	3.140&	\underline{1.847}& \underline{1.848}&	2.323&	2.054\\
\bottomrule
\end{tabular}
\end{table}

\subsubsection{{Conet} Initialization.} We aim to study the impact of {Conet} initialization on model performance. As mentioned in Section \ref{sec:instance}, we create two learnable vectors $\mathbf{v}^{\ell}_b, \mathbf{v}^{\ell}_f$ for Conets. 
We consider three strategies to initialize $\mathbf{v}^{\ell}_b, \mathbf{v}^{\ell}_f$:  
(i) $avg$: with scalar ones; 
(ii) $norm$: with standard normal distribution; 
(iii) $uniform$: with random distribution between 0 and 1.
From Table \ref{table:init}, we observe three strategies perform similarly in most cases, showing stable performance. We also notice $uniform$ and $avg$ initialization performs better than $norm$, which signifies \textit{Dish-TS} and the hidden distribution may be better learned when not using $norm$ initialization.


\subsubsection{Computational Consumption.} We record the extra memory consumption of \textit{Dish-TS}.
As shown in Appendix \ref{sec:app_B.3}, our simple instance of \textit{Dish-TS} (referred in Section \ref{sec:instance}) only causes extra 4MiB (or less) memory consumption, which can be ignored in real-world applications.

\begin{figure}[h]
\centering
\includegraphics[width=0.81\linewidth]{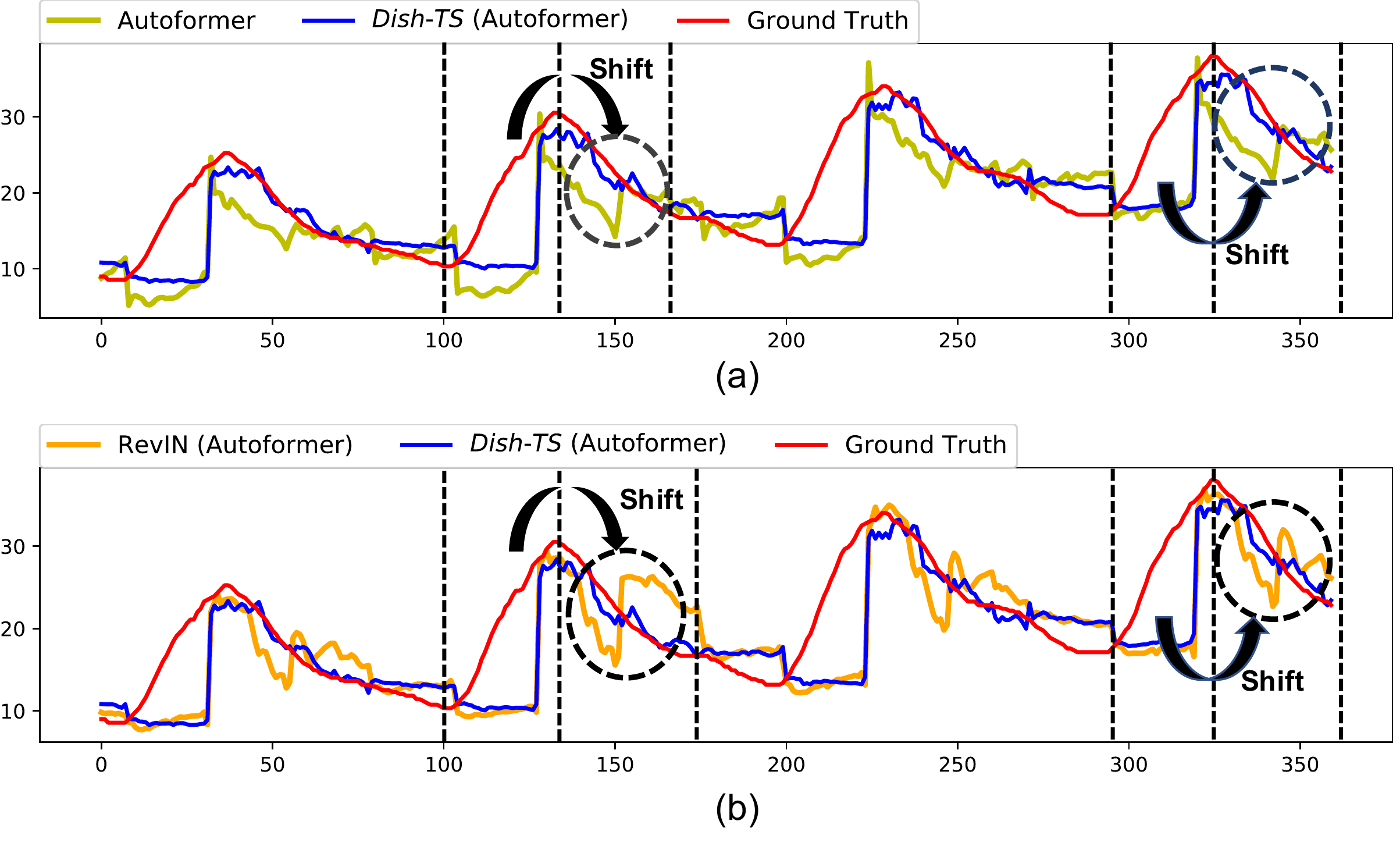}
\caption{Visualizations of backbone (Autoformer), RevIN, and \textit{Dish-TS}, where we highlight results when distribution (series trend) largely changes. 
}
\label{figure:vis}
\end{figure}

\subsubsection{Visualizations.} 
We compare predictions of base model and \textit{Dish-TS} in Figure \ref{figure:vis}(a), predictions of RevIN \cite{kim2022reversible} and \textit{Dish-TS} in Figure \ref{figure:vis}(b). We easily observe when series trend largely changes (could be regarded as the distribution largely changes), both backbone model (Autoformer) and RevIN cannot acquire accurate predictions (in black circles). In contrast, our \textit{Dish-TS} can still make correct forecasting.
We show more visualizations in Appendix \ref{sec:app_B.4}.

\section{Conclusion Remarks}
In this paper, we systematically summarize the distribution shift in time series forecasting as \textit{intra-space shift} and \textit{inter-space shift}. We propose a general paradigm, \textit{Dish-TS} to better alleviate the two shift. To demonstrate the effectiveness, we provide a most simple and intuitive instance of \textit{Dish-TS} along with a prior knowledge-induced training strategy, to couple with state-of-the-art models for better forecasting. We conduct extensive experiments on several datasets and the results demonstrate a very significant improvement over backbone models. We hope this general paradigm together with such an effective instance of \textit{Dish-TS} design can facilitate more future research on distribution shift in time series.

\newpage



\appendix

\begin{table*}[h]
\centering
\footnotesize
\caption{Reported metrics are scaled in experiments for readability.}
\begin{tabular}{c|cc|cc|cc|cc|cc}
\toprule
Dataset& \multicolumn{2}{c}{Electricity} & \multicolumn{2}{c}{ETTh1} & \multicolumn{2}{c}{ETTm2}& \multicolumn{2}{c}{Weather} & \multicolumn{2}{c}{Illness} \\
Metric& MSE & MAE& MSE & MAE& MSE & MAE& MSE & MAE& MSE & MAE\\
\midrule
Uni-variate& $1e^{-6}$&$1e^{-2}$& $1e^{-1}$& $1e^{-1}$ &  $1e^{-1}$&$1e^{-1}$& $1e^{-2}$&$1e^{-1}$ &$1e^{-11}$&$1e^{-5}$ \\
Multi-variate& $1e^{-8}$&$1e^{-3}$& $1e^{-1}$& $1e^{-1}$ &  $1e^{-1}$&$1e^{-1}$& $1e^{-3}$&$1e^{-1}$ &$1e^{-11}$&$1e^{-5}$ \\
\bottomrule
\end{tabular}
\label{table:norm_metric}
\end{table*}

\section{More Experimental Details}

\subsection{More Dataset Details } \label{sec:app_A.1}
We conduct our experiments on the following five real-world datasets:
\begin{itemize}
    \item \textbf{\textit{Electricity}}\footnote{\url{https://archive.ics.uci.edu/ml/datasets/ElectricityLoadDiagrams20112014}} dataset collects the electricity consumption (Kwh) of 321 clients, and raw data is pre-processed following \cite{xu2021autoformer}.
    \item \textit{ETT}\footnote{\url{https://github.com/zhouhaoyi/ETDataset}} datasets contain the data of electricity transformers temperatures of 2-years data collected from two counties in China. The experiments are on two granularity versions of raw data, namely \textbf{\textit{ETTh1}} dataset (1-hour-level) and \textbf{\textit{ETTm2}} dataset (15-minutes-level).
    \item \textbf{\textit{Weather}}\footnote{\url{https://www.bgc-jena.mpg.de/wetter/}} dataset is recorded in ten-minute level in 2020, which contains 21 meteorological indicators, such as air humidity, etc.
    \item \textbf{\textit{Illness}}\footnote{\url{https://gis.cdc.gov/grasp/fluview/fluportaldashboard.html}} dataset has the weekly recorded influenza-like illness (ILI) patients data from Centers for Disease Control and Prevention of the United States between 2002 and 2021, which describes the ratio of patients seen with and the total number of the patients.
\end{itemize}
For data split, we follow \cite{zhou2021informer} and split data into train/validation/test set by the ratio 6:2:2 towards \textit{ETTm1} and \textit{ETTh2} datasets. We follow \cite{xu2021autoformer} to preprocess data and split data by the ratio of 7:1:2 in \textit{Electricity}, \textit{Weather} and \textit{Illness} datasets.

\subsection{More Evaluation Details} \label{sec:app_A.2}
To directly reflects the shift in time series, all experiments and evaluations are conducted on original data without any data normalization or data scaling, which is different from experimental settings of \cite{zhou2021informer,xu2021autoformer} which use z-score normalization to process data before training and evaluation.

In evaluation, we evaluate the time series forecasting performances on the mean squared error (MSE) and mean absolute error (MAE). Note that our experiments and evaluations are on original data; thus the reported metrics are scaled for readability.
Specifically, in univariate forecasting, MSE and MAE is scaled by $1e^{-6}, 1e^{-2}$ on \textit{Electricity}, $1e^{-2}, 1e^{-1}$ on \textit{Weather}, $1e^{-11}, 1e^{-5}$ on \textit{Illness}, and $1e^{-1}, 1e^{-1}$ on all \textit{ETT} datasets.In multivariate forecasting, MSE and MAE is  scaled by $1e^{-8}, 1e^{-3}$ on \textit{Electricity}, $1e^{-3}, 1e^{-1}$ on \textit{Weather}, $1e^{-11}, 1e^{-5}$ on \textit{Illness}, and $1e^{-1}, 1e^{-1}$ on \textit{ETT} datasets. This setting is summarized in Table \ref{table:norm_metric}.

\begin{table*}[h]
\centering
\footnotesize
\caption{Overall performance of univariate time series forecasting on Illness dataset. Due to the limitation of dataset length, we set the lookback and horizon length from 24 to 96.}
\begin{tabular}{c|c|cccc|cccc|cccc}
\toprule
\multicolumn{2}{c}{Method}& \multicolumn{2}{c}{Informer}&\multicolumn{2}{c}{+\textit{Dish-TS}}& \multicolumn{2}{c}{Autoformer}&\multicolumn{2}{c}{+\textit{Dish-TS}}& 
\multicolumn{2}{c}{N-BEATS}&   \multicolumn{2}{c}{+\textit{Dish-TS}}\\
\cmidrule(r){1-2} \cmidrule(r){3-4} \cmidrule(r){5-6} \cmidrule(r){7-8} \cmidrule(r){9-10} \cmidrule(r){11-12} \cmidrule(r){13-14} 
\multicolumn{2}{c}{Metric}&
MSE& MAE&  MSE& MAE&
MSE& MAE&  MSE& MAE&
MSE& MAE&  MSE& MAE\\
\midrule 
\multirow{5}{*}{ \rotatebox{90}{Illness} }
&24&1.405 & 1.159 & \textbf{0.050} & \textbf{0.154}&0.066 & 0.208 & \textbf{0.065} & \textbf{0.199}&0.048 & 0.167 & \textbf{0.046} & \textbf{0.166}\\
&36&1.390 & 1.154 & \textbf{0.036} & \textbf{0.149}&0.050 & 0.183 & \textbf{0.040} & \textbf{0.158}&0.062 & 0.191 & \textbf{0.040} & \textbf{0.155}\\
&48&1.383 & 1.151 & \textbf{0.040} & \textbf{0.169}&0.064 & 0.210 & \textbf{0.054} & \textbf{0.189}&0.053 & 0.174 & \textbf{0.035} & \textbf{0.152}\\
&60&1.398 & 1.158 & \textbf{0.038} & \textbf{0.165}&0.075 & 0.229 & \textbf{0.055} & \textbf{0.195}&0.071 & 0.219 & \textbf{0.039} & \textbf{0.163}\\
&96&1.404 & 1.163 & \textbf{0.090} & \textbf{0.258}&0.148 & 0.336 & \textbf{0.081} & \textbf{0.242}&0.067 & {0.205} & \textbf{0.056} & \textbf{0.201}\\
\bottomrule
\end{tabular}
\label{table:illness_S}
\end{table*}

\begin{table*}[h]
\centering
\footnotesize
\caption{Overall performance of multivariate time series forecasting on Illness dataset.}
\begin{tabular}{c|c|cccc|cccc|cccc}
\toprule
\multicolumn{2}{c}{Method}& \multicolumn{2}{c}{Informer}&\multicolumn{2}{c}{+\textit{Dish-TS}}& \multicolumn{2}{c}{Autoformer}&\multicolumn{2}{c}{+\textit{Dish-TS}}& 
\multicolumn{2}{c}{N-BEATS}&   \multicolumn{2}{c}{+\textit{Dish-TS}}\\
\cmidrule(r){1-2} \cmidrule(r){3-4} \cmidrule(r){5-6} \cmidrule(r){7-8} \cmidrule(r){9-10} \cmidrule(r){11-12} \cmidrule(r){13-14} 
\multicolumn{2}{c}{Metric}&
MSE& MAE&  MSE& MAE&
MSE& MAE&  MSE& MAE&
MSE& MAE&  MSE& MAE\\
\midrule 
\multirow{5}{*}{ \rotatebox{90}{Illness} }
&24&2.009 & 1.718 & \textbf{0.062} & \textbf{0.248}&0.096 & 0.349 & \textbf{0.091} & \textbf{0.308}&0.065 & 0.256 & \textbf{0.062} & \textbf{0.248}\\
&36&1.988 & 1.710 & \textbf{0.052} & \textbf{0.244}&0.078 & 0.328 & \textbf{0.057} & \textbf{0.252}&0.057 & 0.260 & \textbf{0.057} & \textbf{0.257}\\
&48&1.979 & 1.708 & \textbf{0.053} & \textbf{0.259}&0.079 & 0.325 & \textbf{0.069} & \textbf{0.297}&0.054 & 0.245 & \textbf{0.053} & \textbf{0.240}\\
&60&1.999 & 1.720 & \textbf{0.060} & \textbf{0.280}&0.107 & 0.377 & \textbf{0.080} & \textbf{0.324}&0.050 & 0.235 & \textbf{0.048} & \textbf{0.235}\\
&96&2.009 & 1.729 & \textbf{0.130} & \textbf{0.410}&0.203 & 0.514 & \textbf{0.107} & \textbf{0.369}&0.086 & 0.334 & \textbf{0.077} & \textbf{0.311}\\
\bottomrule
\end{tabular}
\label{table:illness_M}
\end{table*}

\subsection{More Implementation Details} \label{sec:app_A.3}

In the main experiments of univariate/multivariate time series forecasting, we let the lookback window and the horizon window have the same length, where we gradually prolong the length as $\{ 24, 48, 96, 168,  336 \}$ to accommodate short-term/long-term settings. For \textit{illness} dataset, the length is set as $\{ 24, 36, 48, 69, 96 \}$ due to the length limitation of the dataset.
The length of horizon is further extended to 720 when fixed the length of lookback as 96 in model analysis experiments, in order to study the long time series forecasting (LSTF) problems \cite{zhou2021informer}.

We train all the models using L2 loss and Adam \cite{kingma2014adam} optimizer with learning rate of [1e-4, 1e-3]. We repeat three times for each experiment and report average performance. The batchsize is set as 256 for Informer, 128 for Autoformer and 1024 for N-BEATS, apart from that in \textit{Electricity} dataset, the batchsize is set 64 for Autoformer and Informer. We set the early stop with 7 steps.
We let $\ell$  equal to 1 to reduce extra consumption as much as possible. 
The rate of prior knowledge guidance $\alpha$ is traversed from 0 to 1. 
All the experiments are implemented with PyTorch \cite{paszke2019pytorch} on single NVIDIA RTX 3090 24GB GPU.

\subsection{More Baseline Details.} \label{sec:app_A.4}
As aforementioned, our \textit{Dish-TS} is a general framework that can be integrated into any deep time series forecasting models. To verify the effectiveness, we couple our dual-conet framework with three state-of-the-art models, Informer \cite{zhou2021informer}, Autoformer \cite{xu2021autoformer} and N-BEATS \cite{Oreshkin2020N-BEATS}. In the main results, we compare the performance of the above three models and their \textit{Dish-TS} coupled versions. Mainly we reproduce the model by following their open source github links.

The detailed implementation details are as follows:
\begin{itemize}
    \item  \textbf{Informer.} We use the official open-source code of \cite{zhou2021informer}\footnote{\url{https://github.com/zhouhaoyi/Informer2020}}. For the hyper-parameter settings, we set them same as original paper, including number of heads, dimension of PropSparse attntion, dimension of feed-forward layer, etc. In addition, we remove time embeddings of Informer to align with N-BEATS.
    
    \item  \textbf{Autoformer.} We use the official open-source code of \cite{xu2021autoformer}\footnote{\url{https://github.com/thuml/Autoformer}}. For hyper-parameter settings, we follow the original settings including moving average of Auto-correlation layer, dimension of Auto-Correlation layer, etc. We also remove time embeddings to align with N-BEATS.
    
    \item \textbf{N-BEATS.} We take the notable reproduced codes of N-BEATS\footnote{\url{https://github.com/ElementAI/N-BEATS}}. We carefully choose the parameter settings of N-BEATS and set number of stacks as 3, number of layers as 10 with 256 as layer size. Note that N-BEATS is an univariate time series forecasting model. To align with the input and output of Informer/Autoformer, we transform the multivariate lookback windows from dimension $K\times L\times N$ to $(K\times N)\times L$, where $K$ is batchsize, $L$ is lookback length and $N$ is number of series. The horizon windows adopt the same strategy. In the mean time, for multivatiate forecasting on different datasets, we set $K$ equals to $[1024/N]$ to avoid very large batch size that influences training.

\end{itemize}

To compare fairly, we compare baselines and their \textit{Dish-TS}-coupled versions under the same experimental settings with all same hyperparameters.

\section{More Experimental Results}

\subsection{More Results of Overall Performance} \label{sec:app_B.1}
In this section, we include more experimental results of overall performance on Illness datasets.

\textit{\textbf{Univariate Time Series Forecasting.}} Table \ref{table:illness_S} shows the univariate time series forecasting performance on \textit{Illness}. We notice that Informer still performs very badly, while \textit{Dish-TS} can still achieve better performances in most cases. Also, the improvement is more obvious when the lookback/horizon length is larger except for few cases, which shows our superiority in forecasting.

\textit{\textbf{Multivariate Time Series Forecasting.}}
Table \ref{table:illness_M} shows the multivariate time series forecasting performance on \textit{Illness} dataset. Firstly, we still observe Informer cannot converge well so we don't discuss it more. From the backbone N-BEATS, we observe the improvement is not that significant except when length is 96. However, From the backbone Autoformer, we find the average improvement is very significant, especially when length is 96, the improvement is $0.148 \rightarrow 0.081$, about $45.3\%$. These observations also show that our \textit{Dish-TS} can handle the situations when lookback/horizon grows larger.

\begin{table*}[t]
\centering
\caption{Performance (MSE) comparison with other normalization techniques on univariate time series forecasting. The improvement is with regard to RevIN.}
\small
\begin{tabular}{c|ccc|ccc|ccc|ccc}
\toprule
Datasets & \multicolumn{3}{c}{Electiricity} & \multicolumn{3}{c}{ETTh1} & \multicolumn{3}{c}{ETTm2} & \multicolumn{3}{c}{Weather}  \\
Length & 24 & 96 & 168 & 24 & 96 & 168 & 24 & 96 & 168 & 24 & 96 & 168  \\
\midrule
RevIN & 1.606& 1.769& 1.470&  0.435& 1.088 & 1.066& 
1.193 &  3.059&  4.068&  0.987& 2.131& 2.113\\
\textit{Dish-TS}& 1.535& 1.330& 1.162&  0.344& 0.872 & 0.959& 
0.762 &  2.385&  3.413&  0.800& 1.824& 1.847\\
\midrule
Improve&4.4\%& 24.8\%& 21.0\%& 20.9\%& 19.8\%& 10.1\%& 
36.2\% & 22.0\%  & 16.1\% &  18.9\%&  14.4\%&  12.5\% \\
\bottomrule
\end{tabular}
\label{table:norm_compare_S}
\end{table*}

\begin{table}[t]
\centering
\fontsize{8.2pt}{9.7}\selectfont
\caption{Memory consumption comparison.} \label{table:memory}
\begin{tabular}{c|cc|cc}
\toprule 
Model & Autoformer & +\textit{Dish-TS} & N-BEATS &+\textit{Dish-TS} \\
\midrule
ETTh1  &  6781MiB&  6783MiB & 2503MiB & 2507MiB \\
Weather &  9085MiB & 9089MiB  & 2495MiB & 2497MiB \\
\bottomrule
\end{tabular}
\end{table}

\subsection{Comparison with Normalization Techniques} \label{sec:app_B.2}
In this section, as shown in Table \ref{table:norm_compare_S},we further report the performance comparison of the state-of-the-art normalization technique RevIN \cite{kim2022reversible}) in univariate time series forecasting.  

From the table, we can still easily observe \textit{Dish-TS} can still perform better than RevIN. The improvement towards RevIN is more than 10\% mostly, up to 36\% in one case. Though RevIN can actually help backbone models improve to some extent, our \textit{Dish-TS} can still achieve better performances compared with RevIN, showing much effectiveness. 
A potential reason for the improvement towards RevIN comes from the considerations towards inter-space shift, which is ignored by RevIN.
Based on this result, we notice that not only in multivariate forecasting but also in univariate forecasting, our \textit{Dish-TS} can perform better than existing works, which demonstrates the superiority of our proposed  \textit{Dish-TS}.

\subsection{Computational Consumption}  \label{sec:app_B.3}
We record the extra memory consumption of \textit{Dish-TS}.
As shown in Table \ref{table:memory}, our simple instance of \textit{Dish-TS} (referred in Section \ref{sec:instance}) only causes extra 4MiB (or less) memory consumption, which can be ignored in real-world applications.


\subsection{More Visualization Cases} \label{sec:app_B.4}
\begin{figure*}[!t]
\centering
\includegraphics[width=0.3\linewidth]{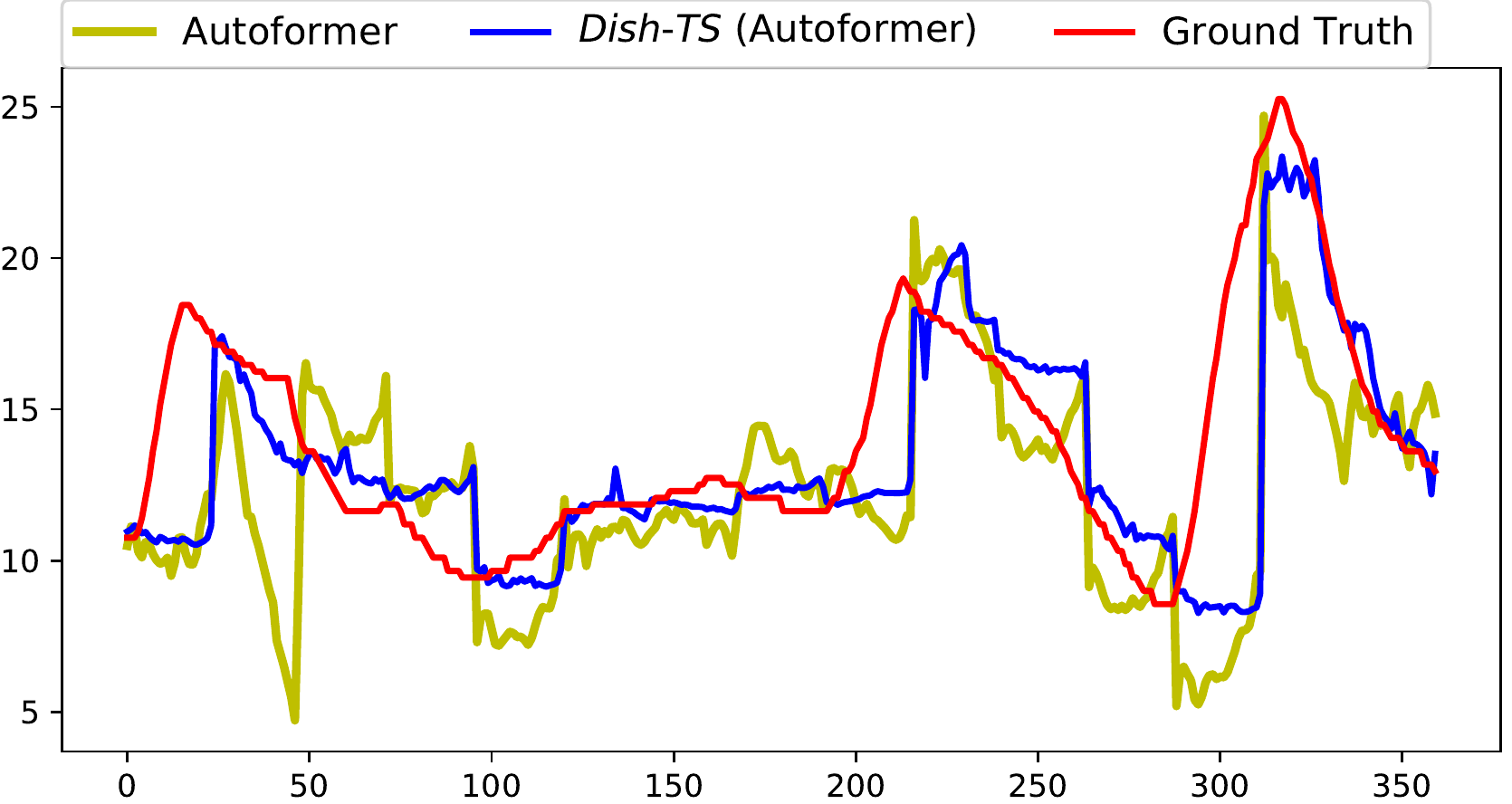}
\includegraphics[width=0.315\linewidth]{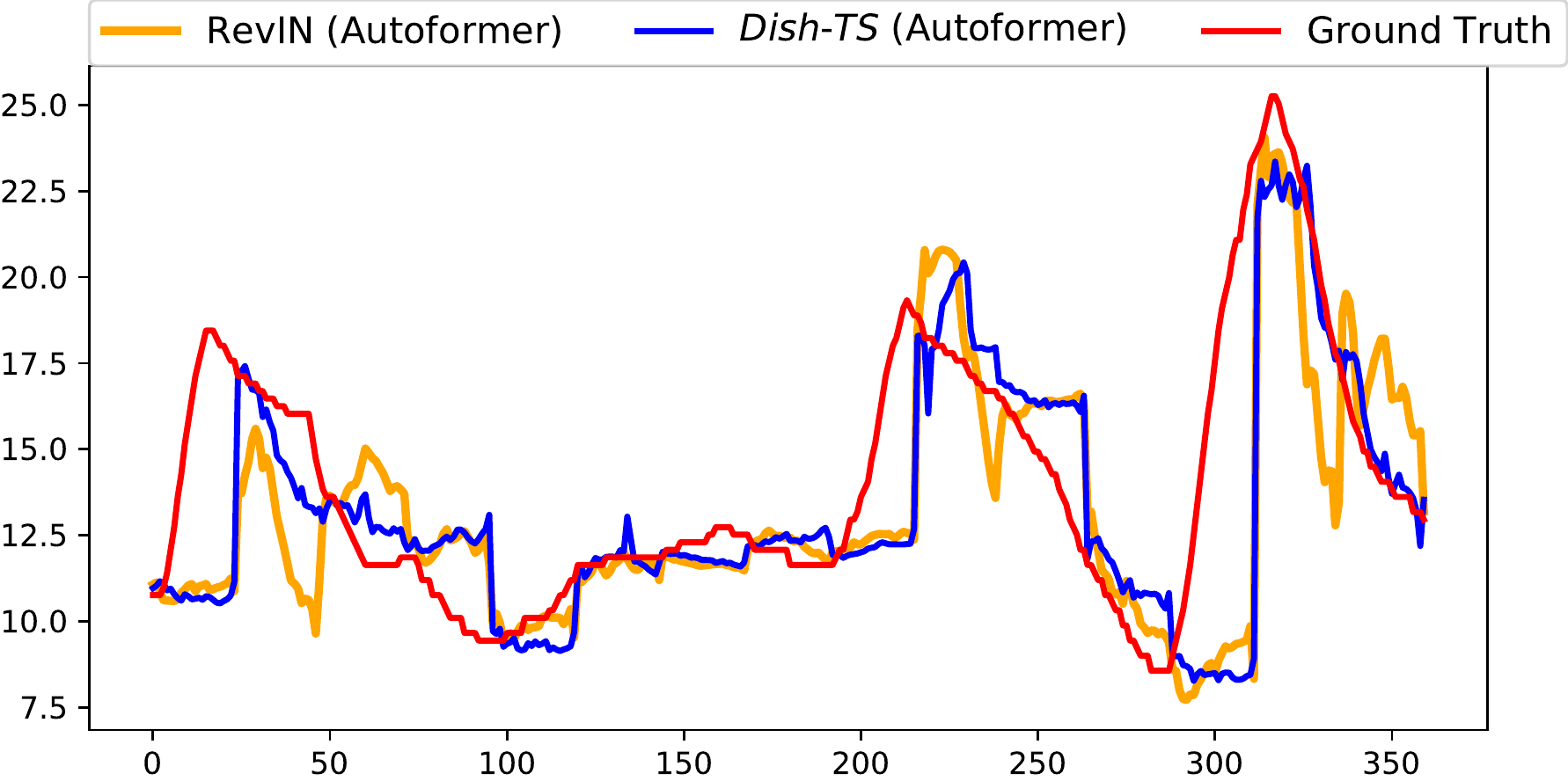}
\caption{Visualizing forecastings on ETTm2 dataset of backbone model (Autoformer), RevIN, and \textit{Dish-TS}.}
\label{fig:vis0}
\end{figure*}

\begin{figure*}[!t]
\centering
\includegraphics[width=0.3\linewidth]{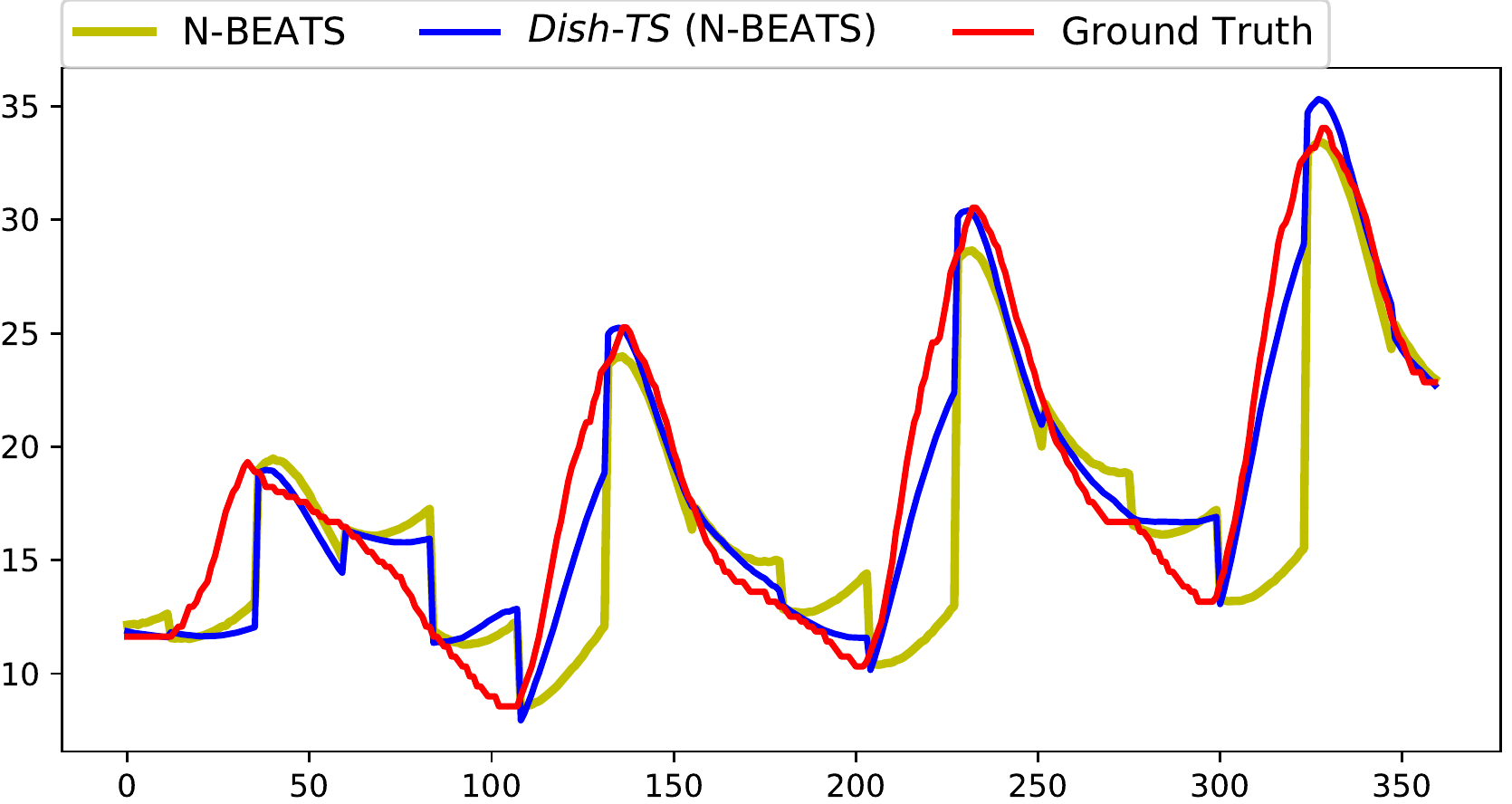}
\includegraphics[width=0.3\linewidth]{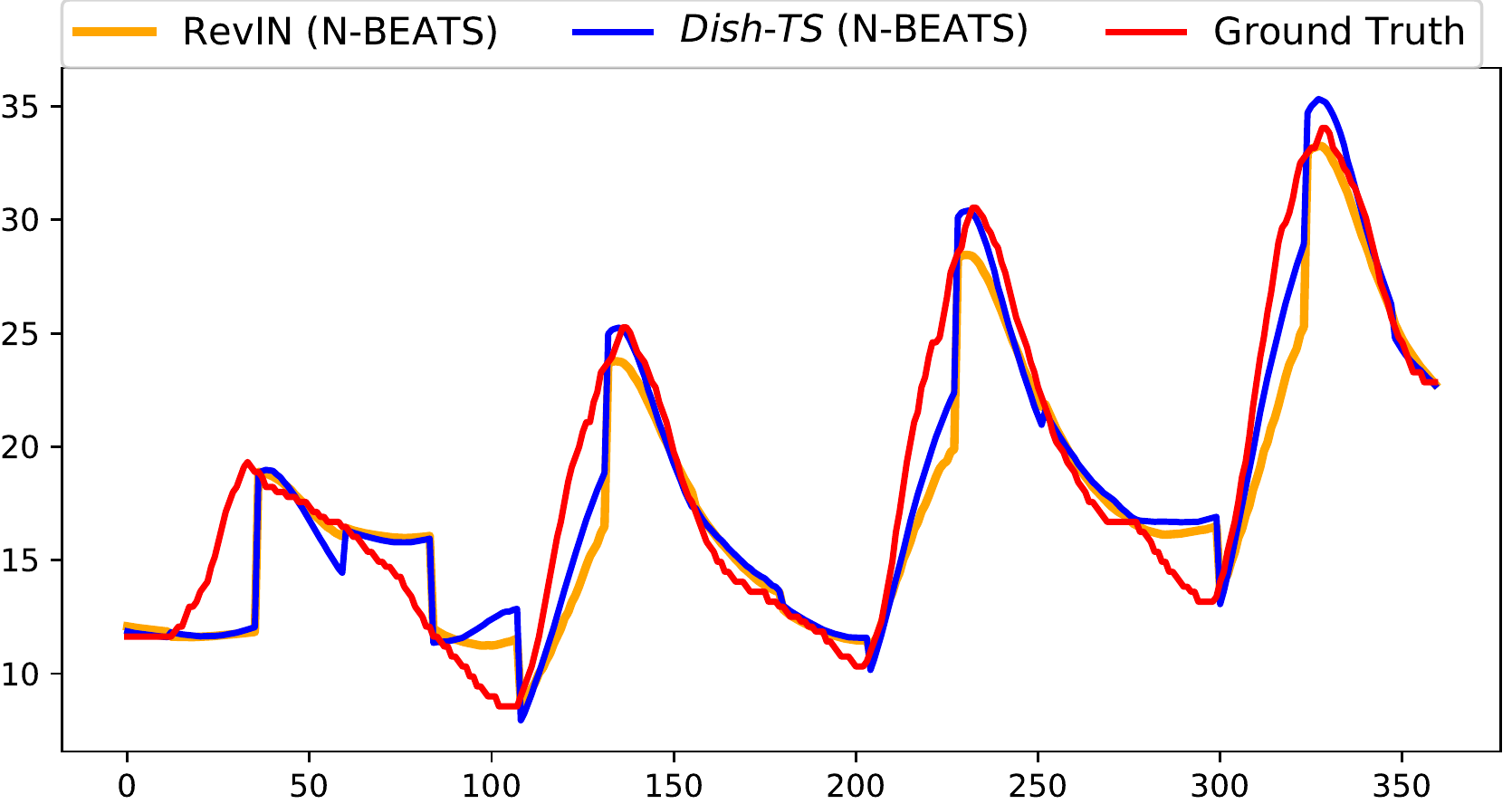}
\caption{Visualizing forecastings on ETTm2 dataset of backbone model (N-BEATS), RevIN, and \textit{Dish-TS}.}
\label{fig:vis1}
\end{figure*}

\begin{figure*}[!t]
\centering
\includegraphics[width=0.3\linewidth]{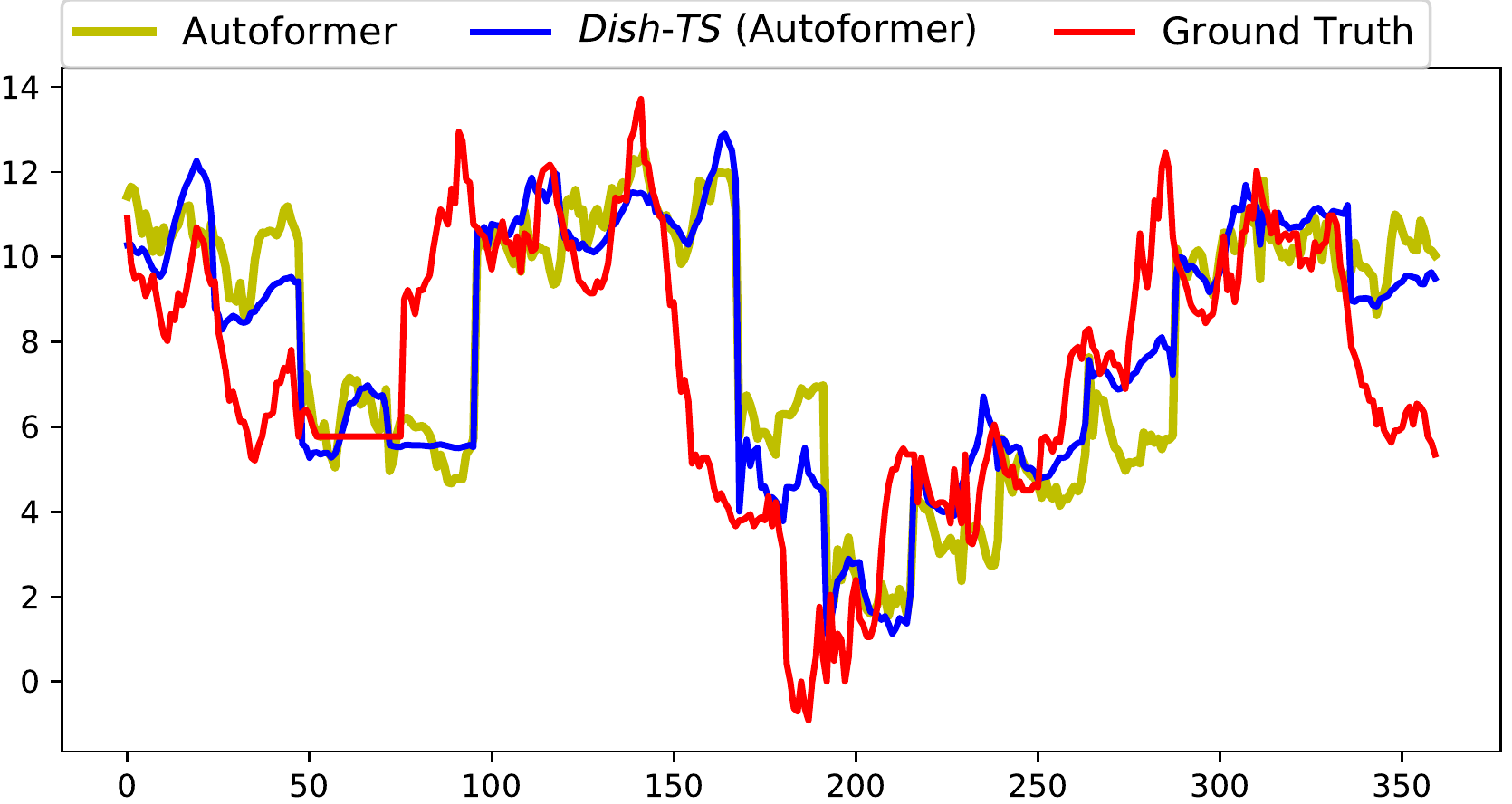}
\includegraphics[width=0.315\linewidth]{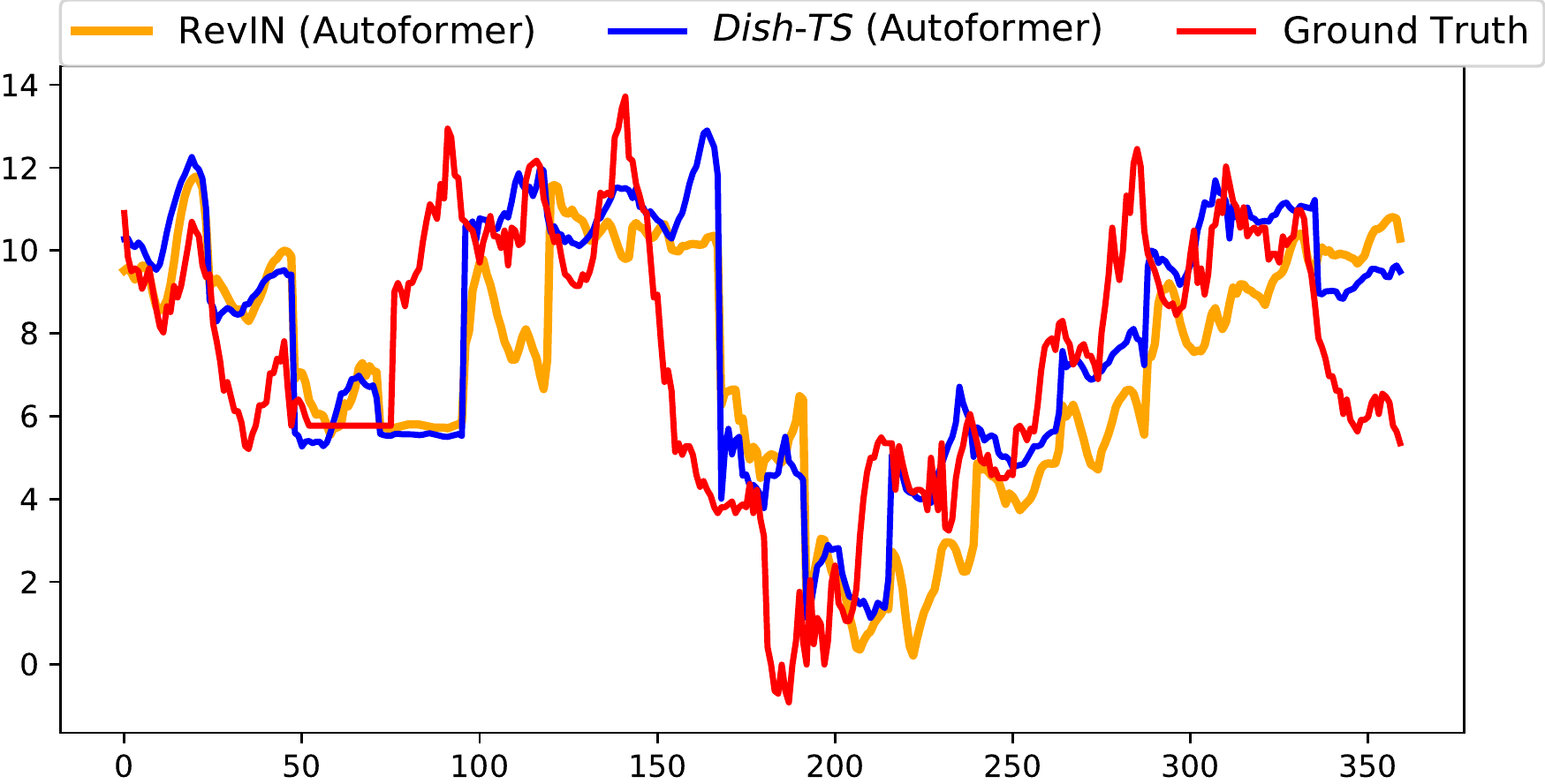}
\caption{Visualizing forecastings on ETTh1 dataset of backbone model (Autoformer), RevIN, and \textit{Dish-TS}.}\label{fig:vis2}
\end{figure*}

\begin{figure*}[!t]
\centering
\includegraphics[width=0.305\linewidth]{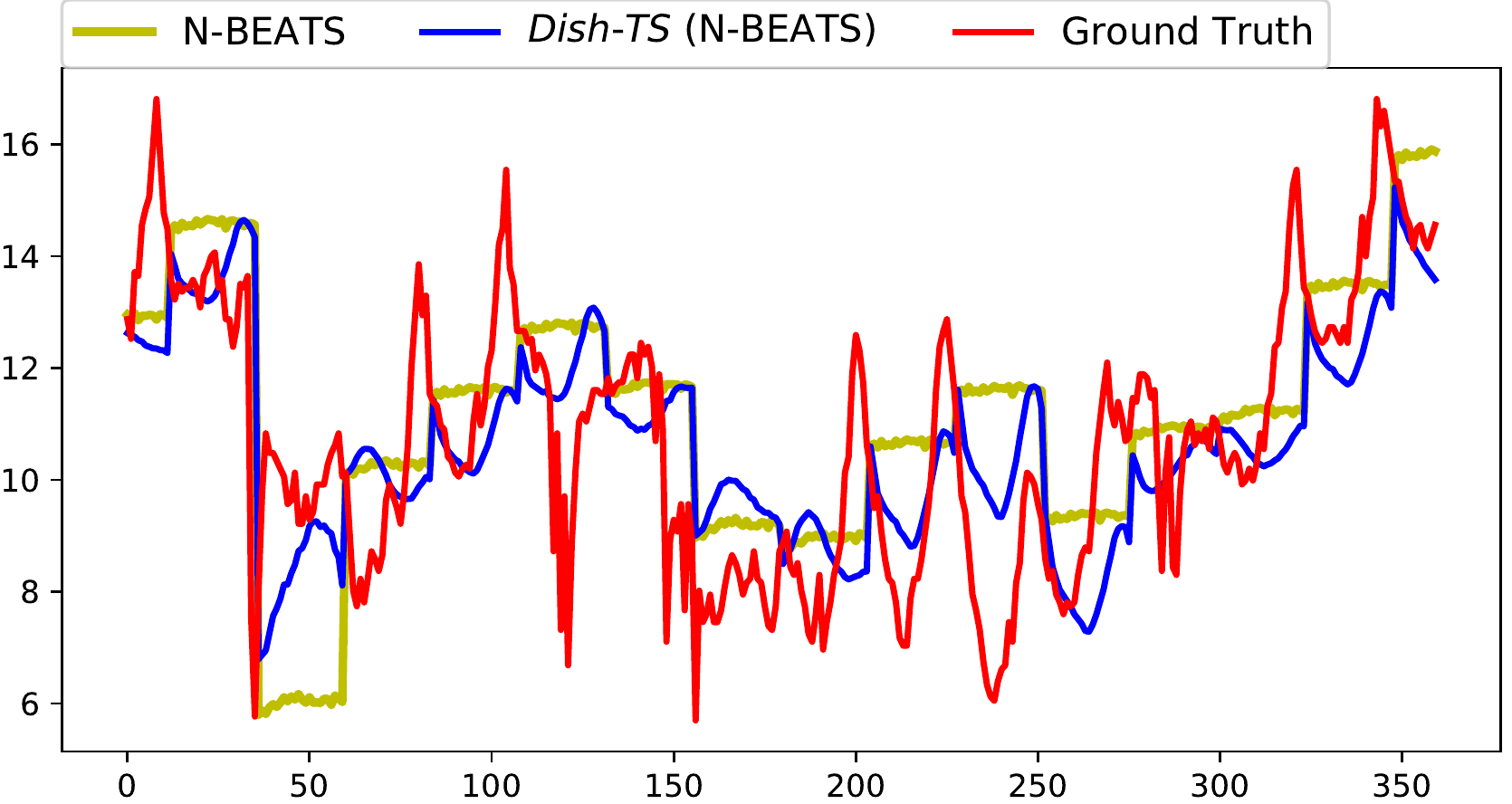}
\includegraphics[width=0.3\linewidth]{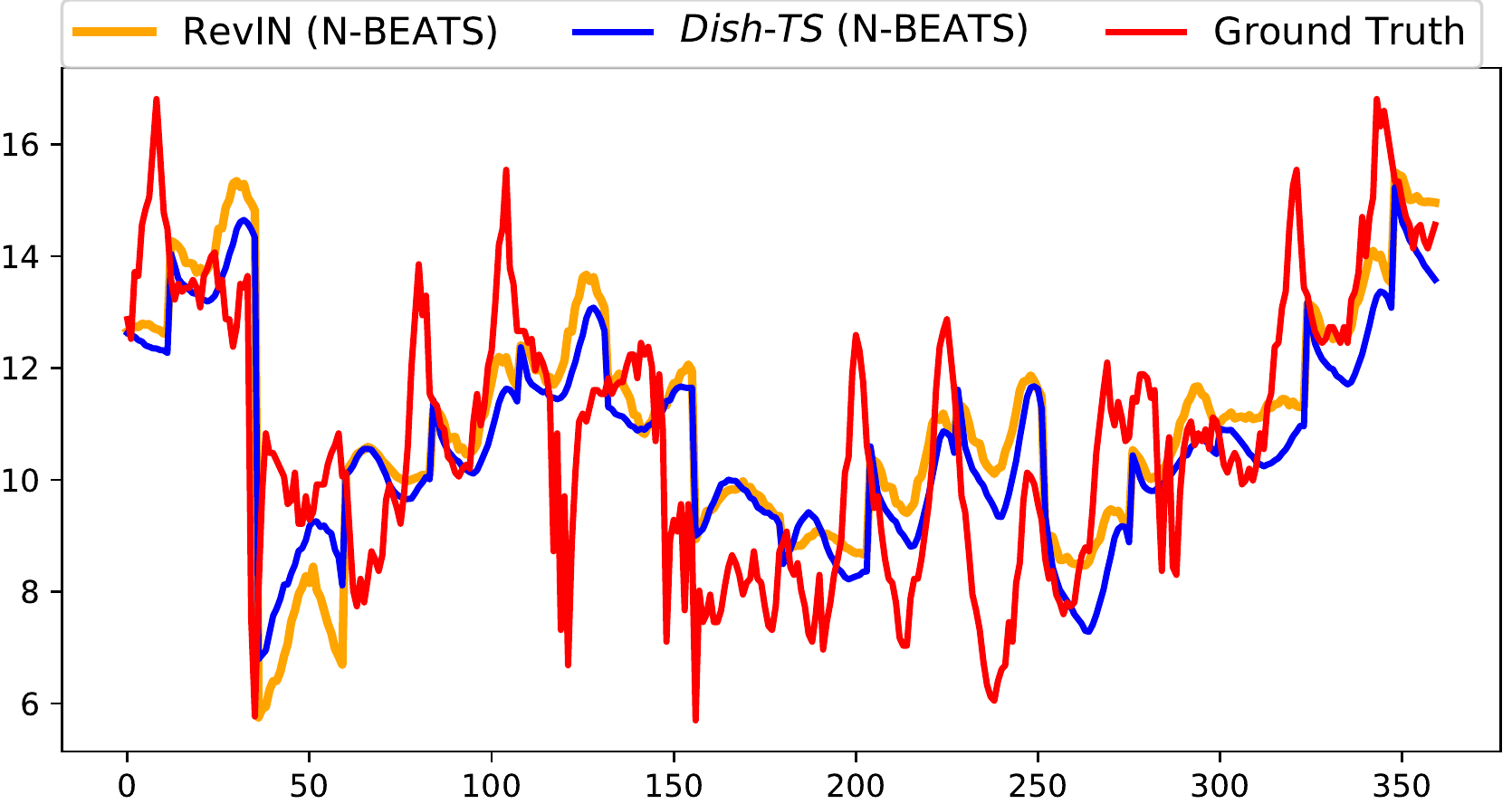}
\caption{Visualizing forecastings on ETTh1 dataset of backbone model (N-BEATS), RevIN, and \textit{Dish-TS}.}
\label{fig:vis3}
\end{figure*}

\begin{figure*}[!t]
\centering
\includegraphics[width=0.3\linewidth]{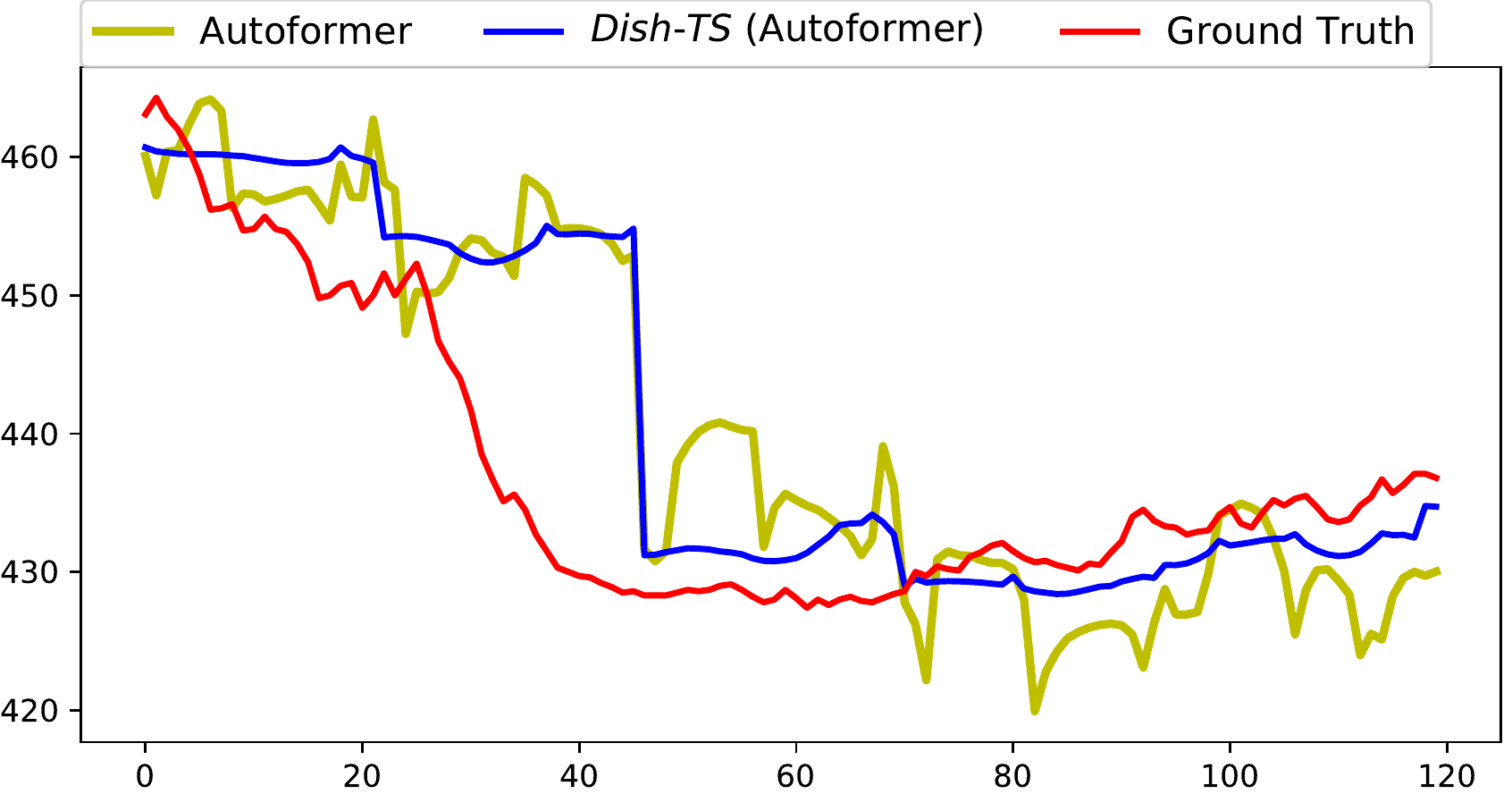}
\includegraphics[width=0.315\linewidth]{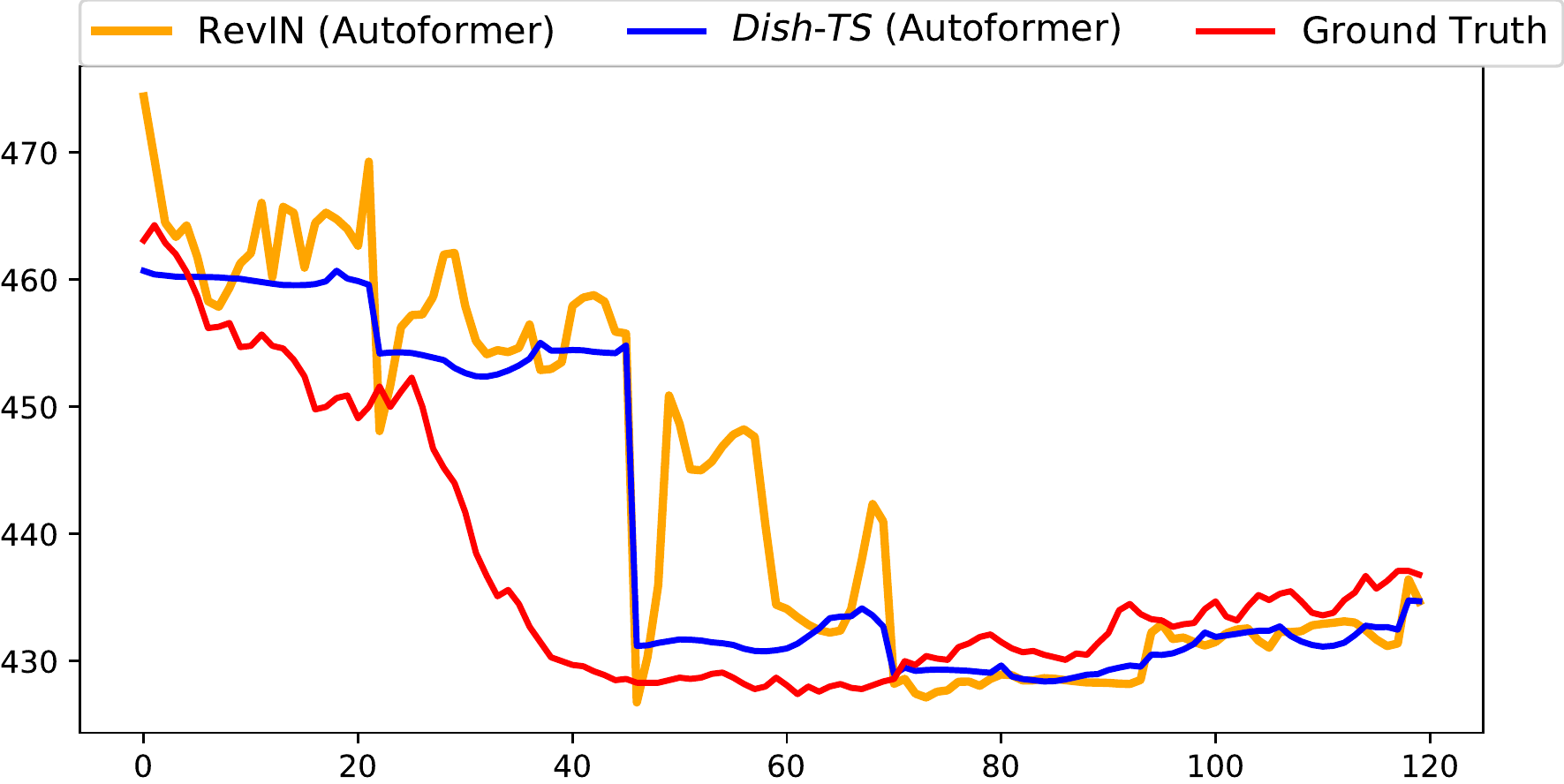}
\caption{Visualizing forecastings on Weather dataset of backbone model (Autoformer), RevIN, and \textit{Dish-TS}.}\label{fig:vis4}
\end{figure*}

\begin{figure*}[!h]
\centering
\includegraphics[width=0.3\linewidth]{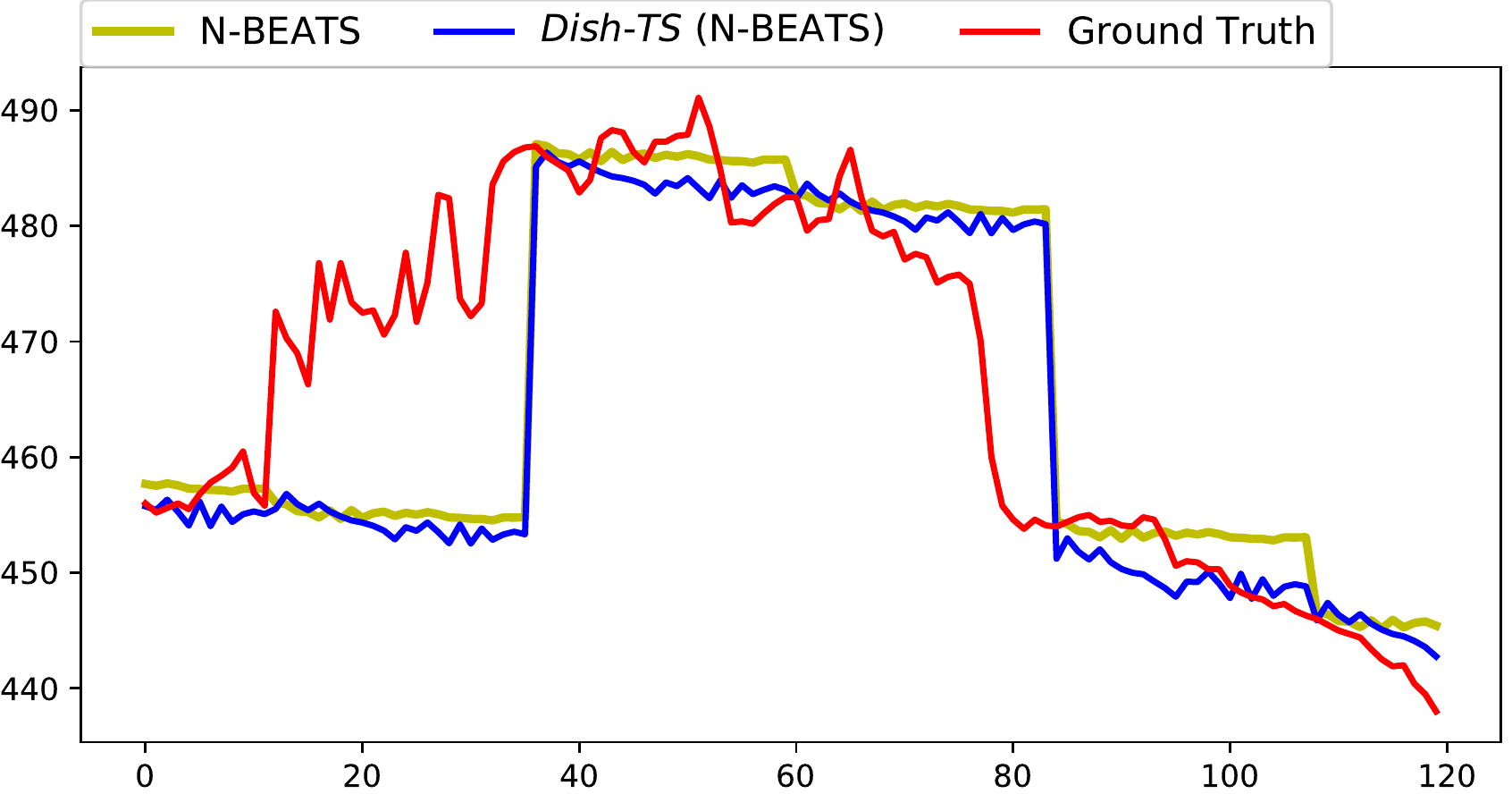}
\includegraphics[width=0.3\linewidth]{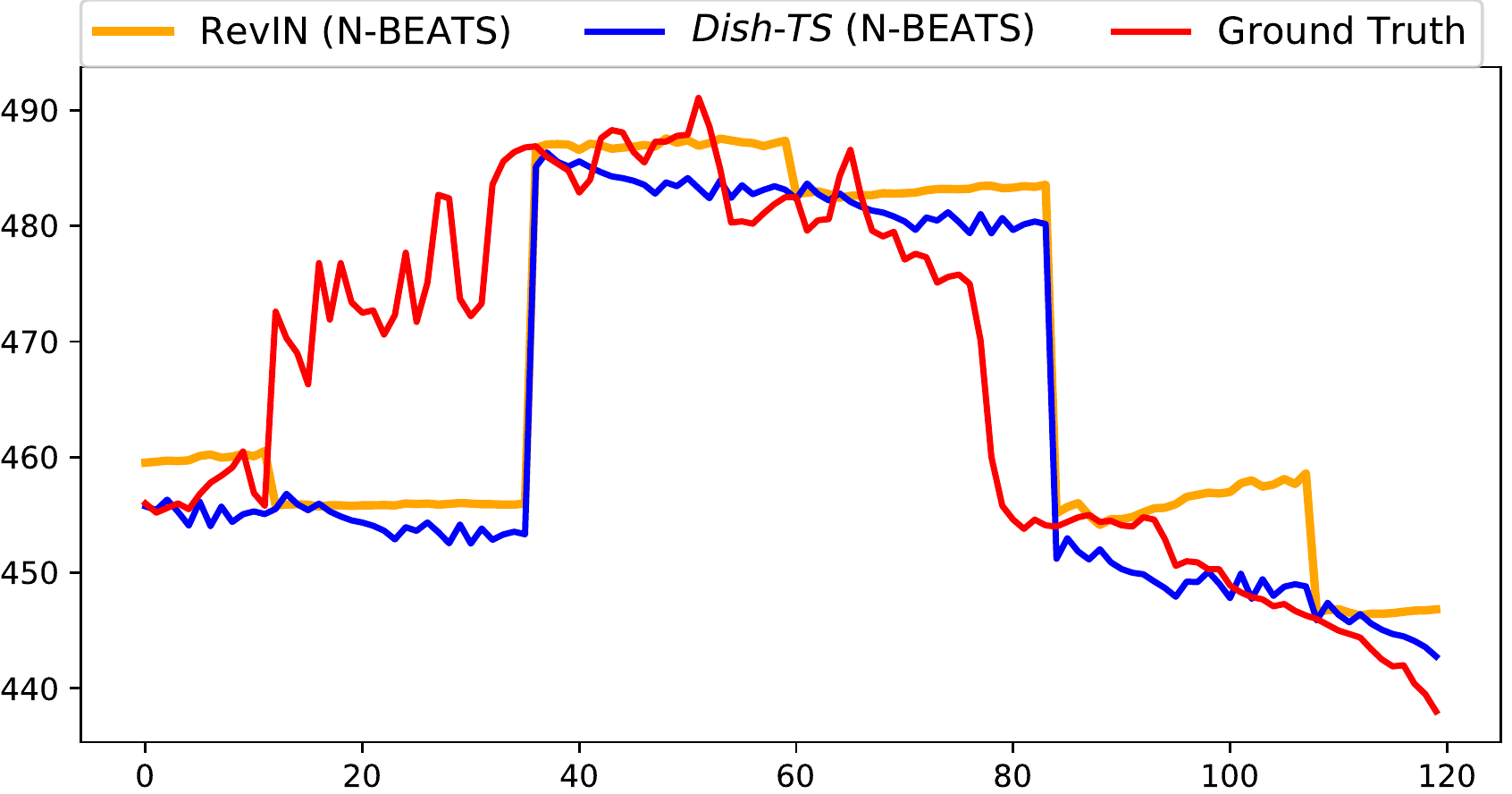}
\caption{Visualizing forecastings on Weather dataset of backbone model (N-BEATS), RevIN, and \textit{Dish-TS}.}\label{fig:vis5}
\end{figure*}

We show more visualization cases on \textit{ETTh1}, \textit{ETTm2}, and \textit{Weather} dataset: Figure  \ref{fig:vis0} and Figure \ref{fig:vis1} shows the forecastings of backbone, RevIN, \textit{Dish-TS} on \textit{ETTm2} dataset, taking Autoformer and N-BEATS as backbones.
Figure \ref{fig:vis2} and Figure \ref{fig:vis3} demonstrate the forecastings of backbone, RevIN, \textit{Dish-TS} on \textit{ETTh1} dataset with Autoformer and  N-BEATS respectively.
Figure \ref{fig:vis4} and Figure \ref{fig:vis5} demonstrate the forecastings of backbone, RevIN, \textit{Dish-TS} on \textit{Weather} dataset with Autoformer and  N-BEATS respectively.

Interestingly, we can consistently notice that when series trend changes dramatically (sudden rise or drop), our \textit{Dish-TS} can help backbone models to acquire more accurate forecasting. This signifies that when series distribution largely changes, \textit{Dish-TS} can be better solution for time series forecasting against distribution shift.
These visualizations show that \textit{Dish-TS} forecasts well in shifted time series.

\subsection{Visualization of Quantified Distribution} \label{sec:app_B.5}
\begin{figure}[!h]
\centering
\includegraphics[width=0.6\linewidth]{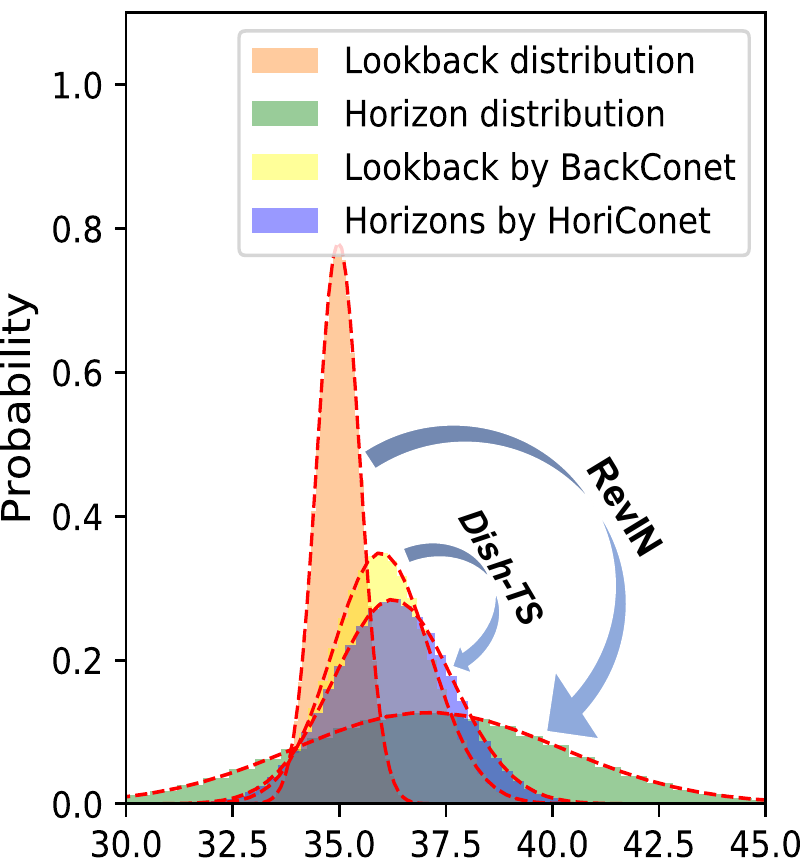}
\caption{Visualization of quantified distribution.}
\label{fig:quantification}
\end{figure}

We further visualize the distribution quantification in Figure \ref{fig:quantification}. Specifically, we first randomly sample 100 lookback windows and 100 horizon windows from test data of \textit{ETTm2} dataset.
We firstly visualize the the distribution of lookback (in orange of Figure \ref{fig:quantification}) and horizon (in green of Figure \ref{fig:quantification}), depicted by mean and std. for the sampled data.
Essentially, the RevIN \cite{kim2022reversible} is to use statistical of lookback distribution to infer the horizon distribution. Thus, the backbone of RevIN needs to go through the transformation of (the blue arrow from orange to green in Figure \ref{fig:quantification}).
We also show the lookback and horizon distribution processed by \textsc{BackConet} and \textsc{HoriConet} of \textit{Dish-TS}, in yellow and in blue of Figure  \ref{fig:quantification} respectively.
Accordingly, we notice the backbone model with \textit{Dish-TS} only needs to conduct mappings (forecasting) of small blue arrow in Figure \ref{fig:quantification}). 
This quantification results show that \textit{Dish-TS} itself models distribution shift and thus leaves less burden to backbone models, which is also an intuitive reason of our significant improvement to RevIN.

\section*{Acknowledgements}
This research was partially supported by the National Science Foundation (NSF) via the grant numbers: 2040950, 2006889, 2045567.

\clearpage

\clearpage

\bibliography{aaai23}

\end{document}